\documentclass{sig-alternate-2013}

\def\b{{\bf b}}
\def\C{{\bf C}}

\def\I{{\bf I}}
\def\R{{\bf R}}
\def\X{{\bf X}}

\def\P{{\bf P}}

\def\S{{\bf S}}

\def\x{{\bf x}}

\def\U{{\bf U}}
\def\u{{\bf u}}
\def\V{{\bf V}}
\def\v{{\bf v}}
\def\W{{\bf W}}

\def\0{{\bf 0}}
\def\1{{\bf 1}}

\def\NM{{\mathcal N}}

\def\ep{\mbox{\boldmath$\epsilon$\unboldmath}}

\usepackage{amsmath}
\usepackage{mathrsfs, euscript}
\usepackage{epsfig,graphicx,subfigure}
\usepackage{multirow}
\usepackage{multicol}
\usepackage{url}
\usepackage{paralist}

\permission{Permission to make digital or hard copies of all or part of this work for personal or classroom use is granted without fee provided that copies are not made or distributed for profit or commercial advantage and that copies bear this notice and the full citation on the first page. Copyrights for components of this work owned by others than ACM must be honored. Abstracting with credit is permitted. To copy otherwise, or republish, to post on servers or to redistribute to lists, requires prior specific permission and/or a fee. Request permissions from Permissions@acm.org.}
\conferenceinfo{KDD'15,}{August 10-13, 2015, Sydney, NSW, Australia.}
\copyrightetc{\copyright~2015 ACM. ISBN \the\acmcopyr}
\crdata{978-1-4503-3664-2/15/08\ ...\$15.00.\\
DOI: http://dx.doi.org/10.1145/2783258.2783273}

\begin{document}
\language0
\lefthyphenmin=2
\righthyphenmin=3
%

\title{Collaborative Deep Learning for Recommender Systems}
%
%
%
%
%

\numberofauthors{3} 
%

\author{
%
%
\alignauthor
Hao Wang\\
       \affaddr{Hong Kong University of Science and Technology}\\
       \email{hwangaz@cse.ust.hk}
\alignauthor
Naiyan Wang\\
       \affaddr{Hong Kong University of Science and Technology}\\
       \email{winsty@gmail.com}
\alignauthor Dit-Yan Yeung\\
       \affaddr{Hong Kong University of Science and Technology}\\
       \email{dyyeung@cse.ust.hk}
}

\maketitle
\begin{abstract}
Collaborative filtering (CF) is a successful approach commonly used by many recommender systems.  Conventional CF-based methods use the ratings given to items by users as the sole source of information for learning to make recommendation.  However, the ratings are often very sparse in many applications, causing CF-based methods to degrade significantly in their recommendation performance.  To address this sparsity problem, auxiliary information such as item content information may be utilized.  Collaborative topic regression (CTR) is an appealing recent method taking this approach which tightly couples the two components that learn from two different sources of information.  Nevertheless, the latent representation learned by CTR may not be very effective when the auxiliary information is very sparse.  To address this problem, we generalize recent advances in deep learning from i.i.d. input to non-i.i.d. (CF-based) input and propose in this paper a hierarchical Bayesian model called collaborative deep learning (CDL), which jointly performs deep representation learning for the content information and collaborative filtering for the ratings (feedback) matrix.  Extensive experiments on three real-world datasets from different domains show that CDL can significantly advance the state of the art.
\end{abstract}


\section{Introduction}\label{sec:intro}

Due to the abundance of choice in many online services, recommender systems (RS) now play an increasingly significant role \cite{DBLP:conf/www/ZhangSLG14}.  For individuals, using RS allows us to make more effective use of information.  Besides, many companies (e.g., Amazon and Netflix) have been using RS extensively to target their customers by recommending products or services. Existing methods for RS can roughly be categorized into three classes \cite{DBLP:journals/kbs/BobadillaOHG13}: content-based methods, collaborative filtering (CF) based methods, and hybrid methods.  Content-based methods \cite{DBLP:conf/icml/Lang95newsweeder} make use of user profiles or product descriptions for recommendation. CF-based methods \cite{DBLP:conf/uai/RendleFGS09,DBLP:conf/nips/SalakhutdinovM07} use the past activities or preferences, such as user ratings on items, without using user or product content information.  Hybrid methods \cite{DBLP:conf/kdd/AgarwalC09,DBLP:conf/ijcai/LiYZ11,DBLP:conf/www/HuCXCGZ13} seek to get the best of both worlds by combining content-based and CF-based methods.

Because of privacy concerns, it is generally more difficult to collect user profiles than past activities.
Nevertheless, CF-based methods do have their limitations.  The prediction accuracy often drops significantly when the ratings are very sparse.  Moreover, they cannot be used for recommending new products which have yet to receive rating information from users.
Consequently, it is inevitable for CF-based methods to exploit auxiliary information and hence hybrid methods have gained popularity in recent years.

According to whether two-way interaction exists between the rating information and auxiliary information, we may further divide hybrid methods into two sub-categories: loosely coupled and tightly coupled methods.
Loosely coupled methods like \cite{DBLP:journals/mta/SevilKDC10} process the auxiliary information once and then use it to provide features for the CF models.  Since information flow is one-way, the rating information cannot provide feedback to guide the extraction of useful features.  For this sub-category, improvement often has to rely on a manual and tedious feature engineering process.  On the contrary, tightly coupled methods like \cite{DBLP:conf/kdd/WangB11} allow two-way interaction.  On one hand, the rating information can guide the learning of features.  On the other hand, the extracted features can further improve the predictive power of the CF models (e.g., based on matrix factorization of the sparse rating matrix).  With two-way interaction, tightly coupled methods can automatically learn features from the auxiliary information and naturally balance the influence of the rating and auxiliary information. This is why tightly coupled methods often outperform loosely coupled ones \cite{DBLP:conf/ijcai/WangCL13}.

Collaborative topic regression (CTR) \cite{DBLP:conf/kdd/WangB11} is a recently proposed tightly coupled method.  It is a probabilistic graphical model that seamlessly integrates a topic model, latent Dirichlet allocation (LDA) \cite{DBLP:journal/jmlr/BleiNJ03}, and a model-based CF method, probabilistic matrix factorization (PMF) \cite{DBLP:conf/nips/SalakhutdinovM07}.
CTR is an appealing method in that it produces promising and interpretable results.  Nevertheless, the latent representation learned is often not effective enough especially when the auxiliary information is very sparse.  It is this representation learning problem that we will focus on in this paper.

On the other hand, deep learning models recently show great potential for learning effective representations and deliver state-of-the-art performance in computer vision \cite{DBLP:conf/nips/WangY13} and natural language processing \cite{DBLP:journals/acl/KalchbrennerGB14,DBLP:journals/ijar/SalakhutdinovH09} applications.  In deep learning models, features are learned in a supervised or unsupervised manner.  Although they are more appealing than shallow models in that the features can be learned automatically (e.g., effective feature representation is learned from text content), they are inferior to shallow models such as CF in capturing and learning the similarity and implicit relationship between items.  This calls for integrating deep learning with CF by performing deep learning collaboratively.

Unfortunately, very few attempts have been made to develop deep learning models for CF. \cite{DBLP:conf/icml/SalakhutdinovMH07} uses restricted Boltzmann machines instead of the conventional matrix factorization formulation to perform CF and \cite{DBLP:conf/icml/GeorgievN13} extends this work by incorporating user-user and item-item correlations. Although these methods involve both deep learning and CF, they actually belong to CF-based methods because they do not incorporate content information like CTR, which is crucial for accurate recommendation. \cite{DBLP:conf/icassp/SainathKSAR13} uses low-rank matrix factorization in the last weight layer of a deep network to significantly reduce the number of model parameters and speed up training, but it is for classification instead of recommendation tasks. On music recommendation, \cite{DBLP:conf/nips/OordDS13,DBLP:conf/mm/WangW14} directly use conventional CNN or deep belief networks (\mbox{DBN}) to assist representation learning for content information, but the deep learning components of their models are deterministic without modeling the noise and hence they are less robust. The models achieve performance boost mainly by loosely coupled methods without exploiting the interaction between content information and ratings. Besides, the CNN is linked directly to the rating matrix, which means the models will perform poorly when the ratings are sparse, as shown in the following experiments.

To address the challenges above, we develop a hierarchical Bayesian model called collaborative deep learning (CDL) as a novel tightly coupled method for RS.  We first present a Bayesian formulation of a deep learning model called stacked denoising autoencoder (SDAE) \cite{DBLP:journals/jmlr/VincentLLBM10}.
With this, we then present our CDL model which tightly couples deep representation learning for the content information and collaborative filtering for the ratings (feedback) matrix, allowing two-way interaction between the two.  Experiments show that CDL significantly outperforms the state of the art.  Note that although we present CDL as using SDAE for its feature learning component, CDL is actually a more general framework which can also admit other deep learning models such as deep Boltzmann machines \cite{DBLP:journals/jmlr/SalakhutdinovH09}, recurrent neural networks \cite{DBLP:conf/icml/GravesFGS06}, and convolutional neural networks \cite{DBLP:conf/nips/KrizhevskySH12}.

The main contribution of this paper is summarized below:
\begin{compactitem}
\item By performing deep learning collaboratively, CDL can simultaneously extract an effective deep feature representation from content and capture the similarity and implicit relationship between items (and users). The learned representation may also be used for tasks other than recommendation.
\item Unlike previous deep learning models which use simple target like classification \cite{DBLP:journals/acl/KalchbrennerGB14} and reconstruction \cite{DBLP:journals/jmlr/VincentLLBM10}, we propose to use CF as a more complex target in a probabilistic framework.
\item Besides the algorithm for attaining maximum a posteriori (MAP) estimates, we also derive a sampling-based algorithm for the Bayesian treatment of CDL, which, interestingly, turns out to be a Bayesian generalized version of back-propagation.
\item To the best of our knowledge, CDL is the first hierarchical Bayesian model to bridge the gap between state-of-the-art deep learning models and RS.
Besides, due to its Bayesian nature, CDL can be easily extended to incorporate other auxiliary information to further boost the performance.
\item Extensive experiments on three real-world datasets from different domains show that CDL can significantly advance the state of the art.
\end{compactitem}

\section{Notation and Problem Formulation}

Similar to the work in \cite{DBLP:conf/kdd/WangB11}, the recommendation task considered in this paper takes implicit feedback \cite{DBLP:conf/icdm/HuKV08} as the training and test data.  The entire collection of $J$ items (articles or movies) is represented by a $J$-by-$S$ matrix $\X_c$, where row $j$ is the bag-of-words vector $\X_{c,j*}$ for item $j$ based on a vocabulary of size $S$.  With $I$ users, we define an $I$-by-$J$ binary rating matrix $\R=[\R_{ij}]_{I\times J}$.
For example, in the dataset \emph{citeulike-a} $\R_{ij}=1$ if user $i$ has article $j$ in his or her personal library and $\R_{ij}=0$ otherwise.
Given part of the ratings in $\R$ and the content information $\X_c$, the problem is to predict the other ratings in $\R$. Note that although we focus on movie recommendation (where plots of movies are considered as content information) and article recommendation like \cite{DBLP:conf/kdd/WangB11} in this paper, our model is general enough to handle other recommendation tasks (e.g., tag recommendation).

The matrix $\X_c$ plays the role of clean input to the SDAE while the noise-corrupted matrix, also a $J$-by-$S$ matrix, is denoted by $\X_0$.  The output of layer $l$ of the SDAE is denoted by $\X_l$ which is a $J$-by-$K_l$ matrix.  Similar to $\X_c$, row $j$ of $\X_l$ is denoted by $\X_{l,j*}$.  $\W_l$ and $\b_l$ are the weight matrix and bias vector, respectively, of layer $l$, $\W_{l,*n}$ denotes column $n$ of $\W_l$, and $L$ is the number of layers. For convenience, we use $\W^+$ to denote the collection of all layers of weight matrices and biases. Note that an $L/2$-layer SDAE corresponds to an $L$-layer network.

\begin{figure*}[!tb]
\begin{center}
\subfigure{
\includegraphics[height=3.6cm]{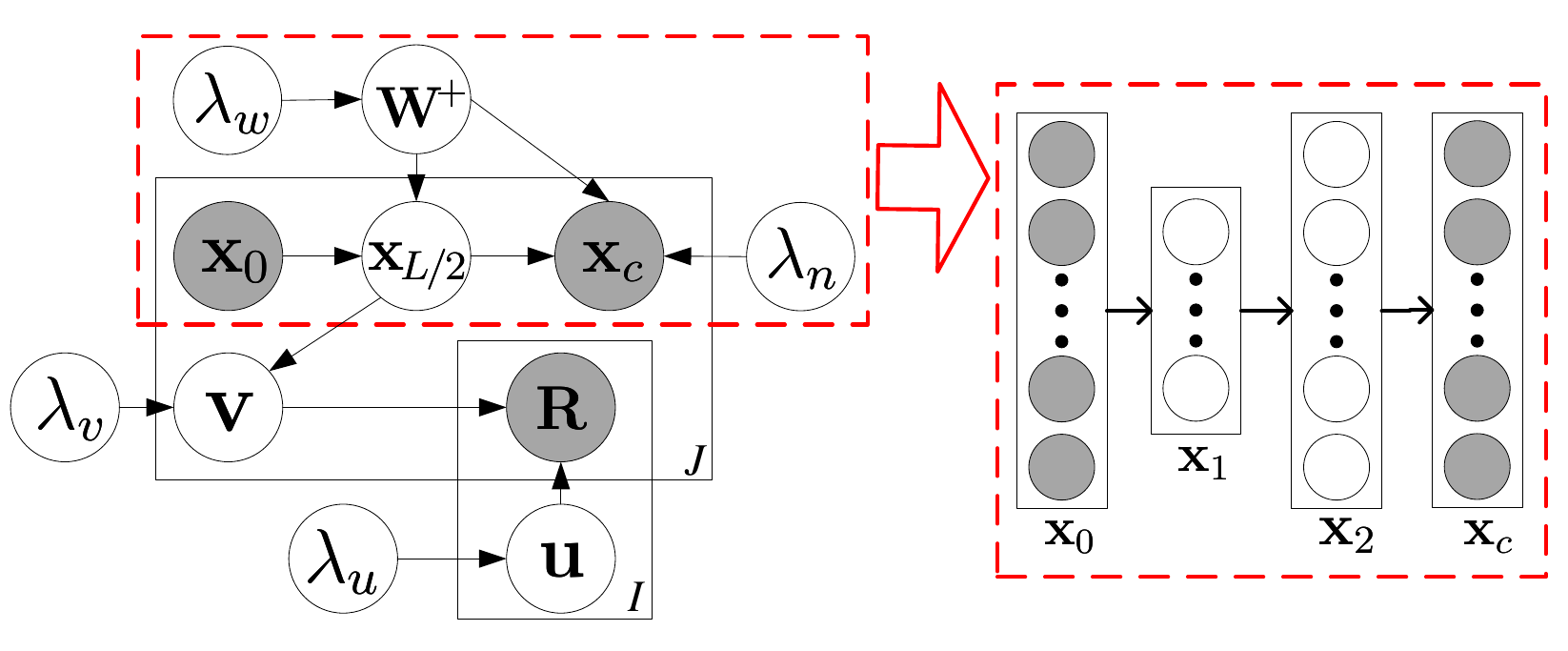}}
\hspace{0.3in}
\subfigure{
\includegraphics[height=3.4cm]{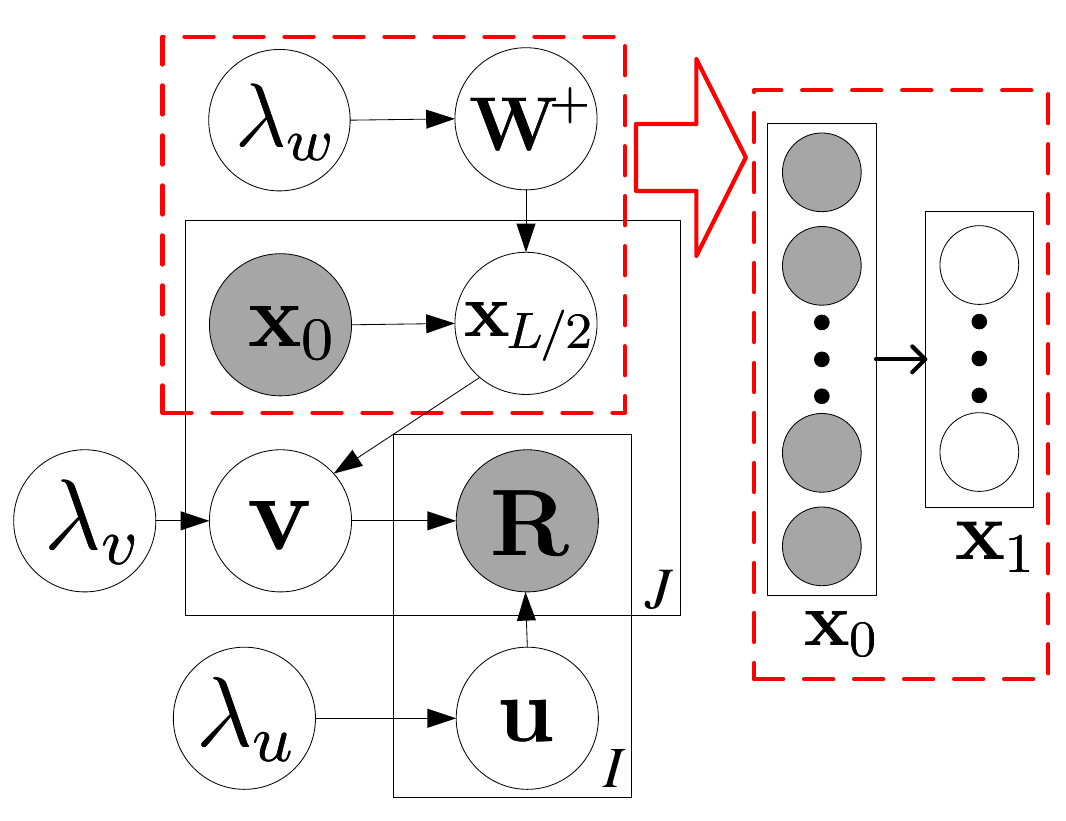}}
\end{center}
\vskip -0.2in
\caption{On the left is the graphical model of CDL. The part inside the dashed rectangle represents an SDAE.  An example SDAE with $L = 2$ is shown. On the right is the graphical model of the degenerated CDL. The part inside the dashed rectangle represents the encoder of an SDAE. An example SDAE with $L=2$ is shown on the right of it. Note that although $L$ is still $2$, the decoder of the SDAE vanishes. To prevent clutter, we omit all variables $\x_l$ except $\x_0$ and $\x_{L/2}$ in the graphical models.
}
\label{fig:cdl_pgm}
\vskip -0.2in
\end{figure*}

\section{Collaborative Deep Learning}
\label{CDL}

We are now ready to present details of our CDL model.  We first briefly review SDAE and give a Bayesian formulation of SDAE.  This is then followed by the presentation of CDL as a hierarchical Bayesian model which tightly integrates the ratings and content information.

\subsection{Stacked Denoising Autoencoders}

\begin{figure}[!tb]
\begin{center}
\vskip -0.15in
\includegraphics[height=3.5cm]{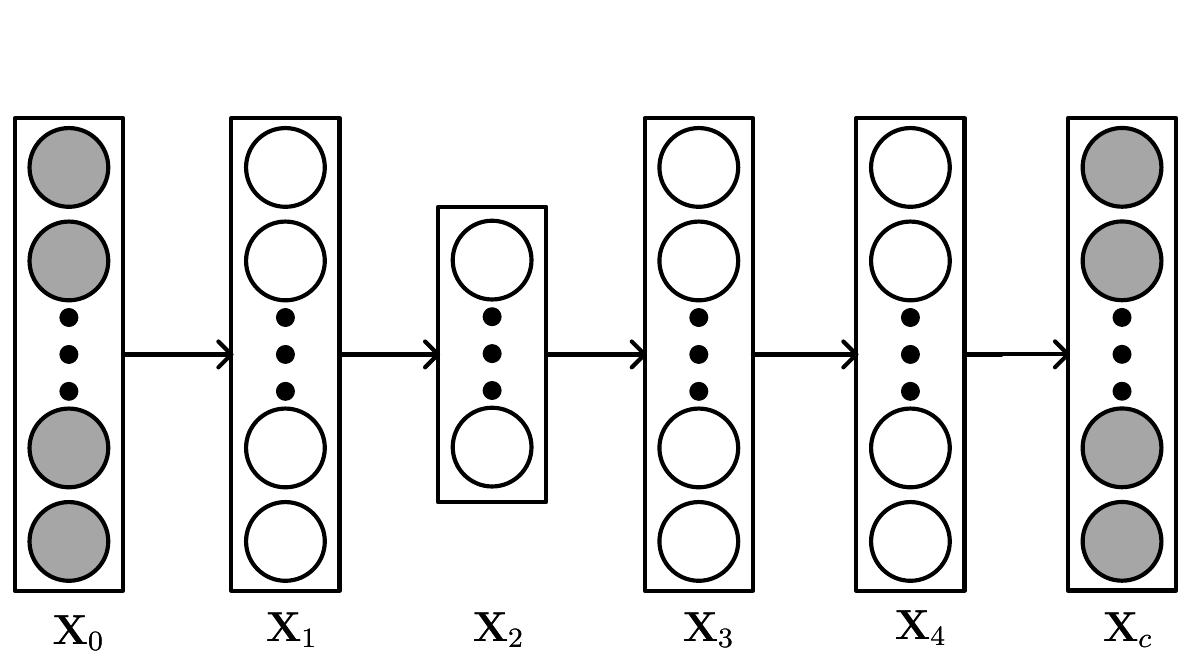}
\end{center}
\vskip -0.2in
\caption{A 2-layer SDAE with $L=4$.}
\label{fig:sdae_ctr}
\vskip -0.2in
\end{figure}

SDAE \cite{DBLP:journals/jmlr/VincentLLBM10} is a feedforward neural network for learning representations (encoding) of the input data by learning to predict the clean input itself in the output, as shown in Figure \ref{fig:sdae_ctr}. Usually the hidden layer in the middle, i.e., $\X_2$ in the figure, is constrained to be a bottleneck and the input layer $\X_0$ is a corrupted version of the clean input data. An SDAE solves the following optimization problem:
\begin{align}
\min\limits_{\{\W_l\},\{\b_l\}} \|\X_c-\X_L\|_F^2+\lambda\sum\limits_l \|\W_l\|_F^2,\nonumber
\end{align}
where $\lambda$ is a regularization parameter and $\|\cdot\|_F$ denotes the Frobenius norm.

\subsection{Generalized Bayesian SDAE}

If we assume that both the clean input $\X_c$ and the corrupted input $\X_0$ are observed, similar to \cite{Bishop2006,DBLP:journals/neco/MacKay92a,DBLP:conf/nips/BengioYAV13,DBLP:conf/icml/ChenXWS12},
we can define the following generative process:
\begin{enumerate}
\item For each layer $l$ of the SDAE network,
\begin{enumerate}
\item For each column $n$ of the weight matrix $\W_l$, draw
\begin{align*}
\W_{l,*n} \sim \NM(\0,\lambda_w^{-1} \I_{K_l}).
\end{align*}
\item Draw the bias vector $\b_l \sim \NM(\0,\lambda_w^{-1} \I_{K_l})$.
\item For each row $j$ of $\X_l$, draw
\begin{align}\label{eq:gaussian}
\X_{l,j*} \sim \NM(\sigma(\X_{l-1,j*}\W_l+\b_l),\lambda_s^{-1} \I_{K_l}).
\end{align}
\end{enumerate}
\item For each item $j$, draw a clean input
\footnote{Note that while generation of the \emph{clean} input $\X_c$ from $\X_L$ is part of the generative process of the Bayesian SDAE, generation of the \emph{noise-corrupted} input $\X_0$ from $\X_c$ is an artificial noise injection process to help the SDAE learn a more robust feature representation.}
\begin{align*}
\X_{c,j*} \sim \NM(\X_{L,j*},\lambda_n^{-1}\I_{J}).
\end{align*}
\end{enumerate}
Note that if $\lambda_s$ goes to infinity, the Gaussian distribution in Equation (\ref{eq:gaussian}) will become a Dirac delta distribution \cite{strichartz2003guide} centered at $\sigma(\X_{l-1,j*}\W_l+\b_l)$, where $\sigma(\cdot)$ is the sigmoid function. The model will degenerate to be a Bayesian formulation of SDAE. That is why we call it generalized SDAE.

Note that the first $L/2$ layers of the network act as an encoder and the last $L/2$ layers act as a decoder.  Maximization of the posterior probability is equivalent to minimization of the reconstruction error with weight decay taken into consideration.

\subsection{Collaborative Deep Learning}

Using the Bayesian SDAE as a component, the generative process of CDL is defined as follows:
\begin{enumerate}
\item For each layer $l$ of the SDAE network,
\begin{enumerate}
\item For each column $n$ of the weight matrix $\W_l$, draw
\begin{displaymath}
\W_{l,*n} \sim \NM(\0,\lambda_w^{-1} \I_{K_l}).
\end{displaymath}
\item Draw the bias vector $\b_l \sim \NM(\0,\lambda_w^{-1} \I_{K_l})$.
\item For each row $j$ of $\X_l$, draw
\begin{displaymath}
\X_{l,j*} \sim \NM(\sigma(\X_{l-1,j*}\W_l+\b_l),\lambda_s^{-1} \I_{K_l}).
\end{displaymath}
\end{enumerate}
\item For each item $j$,
\begin{enumerate}
\item Draw a clean input $\X_{c,j*} \sim \NM(\X_{L,j*},\lambda_n^{-1} \I_{J}$).
\item Draw a latent item offset vector $\ep_j \sim \NM(\0,\lambda_v^{-1}\I_K)$ and then set the latent item vector to be:
\begin{displaymath}
\v_j=\ep_j+\X_{\frac{L}{2},j*}^T.
\end{displaymath}
\end{enumerate}
\item Draw a latent user vector for each user $i$:
\begin{displaymath}
\u_i \sim \NM(\0,\lambda_u^{-1}\I_K).
\end{displaymath}
\item Draw a rating $\R_{ij}$ for each user-item pair $(i,j)$:
\begin{displaymath}
\R_{ij} \sim \NM(\u_i^T\v_j,\C_{ij}^{-1}).
\end{displaymath}
\end{enumerate}
Here $\lambda_w$, $\lambda_n$, $\lambda_u$, $\lambda_s$, and $\lambda_v$ are hyperparameters
and $\C_{ij}$ is a confidence parameter similar to that for CTR ($\C_{ij} = a$ if $\R_{ij}=1$ and $\C_{ij}=b$ otherwise). Note that the middle layer $\X_{L/2}$ serves as a bridge between the ratings and content information. This middle layer, along with the latent offset $\ep_j$, is the key that enables CDL to simultaneously learn an effective feature representation and capture the similarity and (implicit) relationship between items (and users). Similar to the generalized SDAE, for computational efficiency, we can also take $\lambda_s$ to infinity.

The graphical model of CDL when $\lambda_s$ approaches positive infinity is shown in Figure \ref{fig:cdl_pgm}, where, for notational simplicity, we use $\x_0$, $\x_{L/2}$, and $\x_L$ in place of $\X_{0,j*}^T$, $\X_{\frac{L}{2},j*}^T$, and $\X_{L,j*}^T$, respectively.

\subsection{Maximum A Posteriori Estimates}

\begin{figure}[!tb]
\begin{center}
\vskip -0.2in
\includegraphics[height=6.0cm]{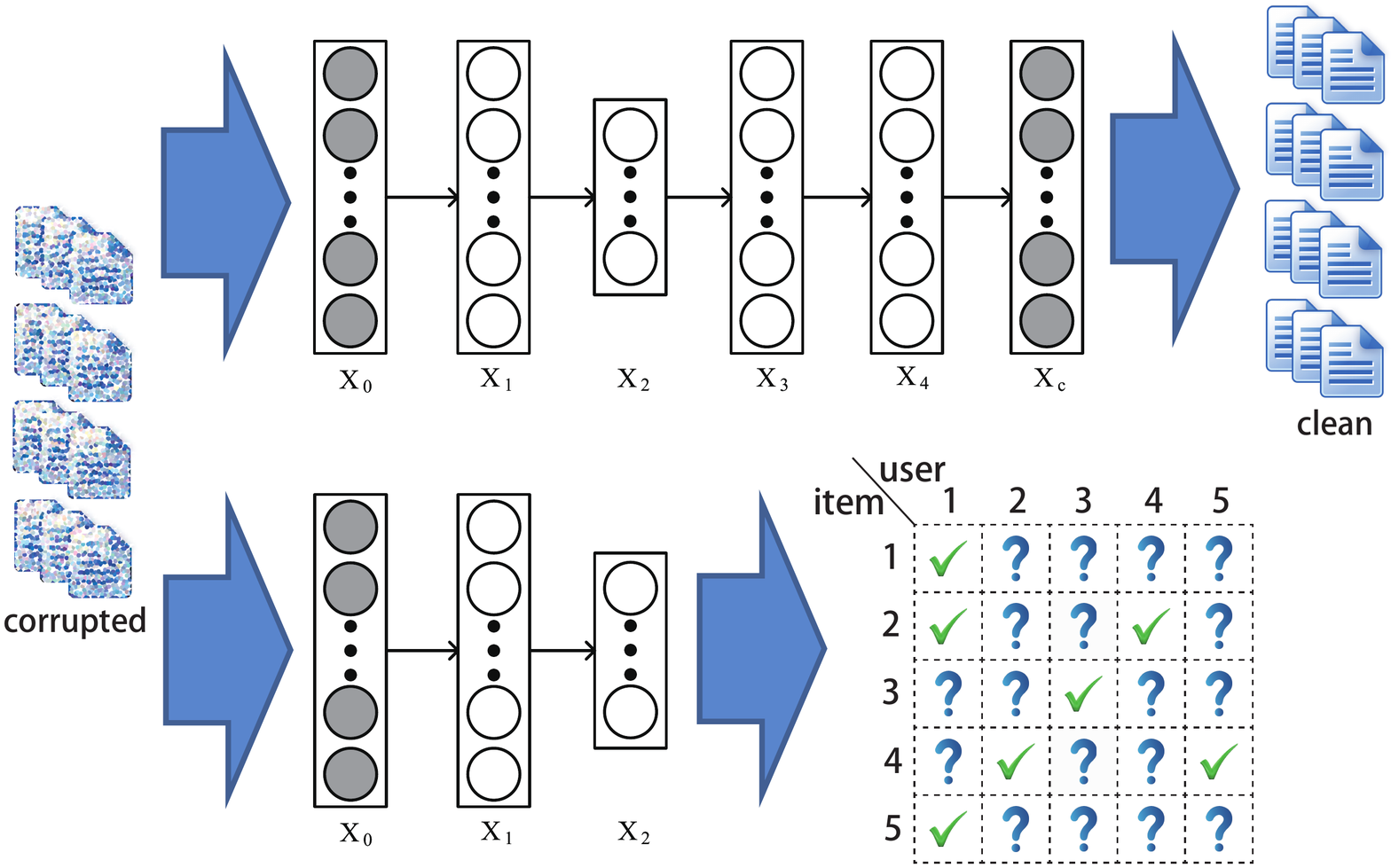}
\end{center}
\vskip -0.4in
\caption{NN representation for degenerated CDL.
}
\label{fig:twonet}
\vskip -0.2in
\end{figure}

Based on the CDL model above, all parameters could be treated as random variables so that fully Bayesian methods such as Markov chain Monte Carlo (MCMC) or variational approximation methods \cite{DBLP:journals/ml/JordanGJS99} may be applied.  However, such treatment typically incurs high computational cost.  Besides, since CTR is our primary baseline for comparison, it would be fair and reasonable to take an approach analogous to that used in CTR.  Consequently, we devise below an EM-style algorithm for obtaining the MAP estimates, as in \cite{DBLP:conf/kdd/WangB11}.

Like in CTR, maximizing the posterior probability is equivalent to maximizing the joint log-likelihood of $\U$, $\V$, $\{\X_l\}$, $\X_c$, $\{\W_l\}$, $\{\b_l\}$, and $\R$ given $\lambda_u$, $\lambda_v$, $\lambda_w$, $\lambda_s$, and $\lambda_n$:
\begin{align}
\mathscr{L}=&-\frac{\lambda_u}{2}\sum\limits_i \|\u_i\|_2^2
-\frac{\lambda_w}{2}\sum\limits_l(\|\W_l\|_F^2+\|\b_l\|_2^2) \nonumber\\
&-\frac{\lambda_v}{2}\sum\limits_j\|\v_j-\X_{\frac{L}{2},j*}^T\|_2^2
-\frac{\lambda_n}{2}\sum\limits_j\|\X_{L,j*}-\X_{c,j*}\|_2^2 \nonumber \\
&-\frac{\lambda_s}{2}\sum\limits_l\sum\limits_j\|\sigma(\X_{l-1,j*}\W_l+\b_l)-\X_{l,j*}\|_2^2 \nonumber \\
&-\sum\limits_{i,j}\frac{\C_{ij}}{2}(\R_{ij}-\u_i^T\v_j)^2. \nonumber
\end{align}\label{eq:Lgen}
If $\lambda_s$ goes to infinity, the likelihood becomes:
\begin{align}\label{eq:L}
\mathscr{L}=&-\frac{\lambda_u}{2}\sum\limits_i \|\u_i\|_2^2
-\frac{\lambda_w}{2}\sum\limits_l(\|\W_l\|_F^2+\|\b_l\|_2^2)\nonumber \\
&-\frac{\lambda_v}{2}\sum\limits_j\|\v_j-f_e(\X_{0,j*},\W^+)^T\|_2^2 \nonumber \\
&-\frac{\lambda_n}{2}\sum\limits_j\|f_r(\X_{0,j*},\W^+)-\X_{c,j*}\|_2^2 \nonumber \\
&-\sum\limits_{i,j}\frac{\C_{ij}}{2}(\R_{ij}-\u_i^T\v_j)^2,
\end{align}
where the encoder function $f_e(\cdot,\W^+)$ takes the corrupted content vector $\X_{0,j*}$ of item $j$ as input and computes the encoding of the item, and the function $f_r(\cdot,\W^+)$ also takes $\X_{0,j*}$ as input, computes the encoding and then the reconstructed content vector of item $j$.
For example, if the number of layers $L=6$, $f_e(\X_{0,j*},\W^+)$ is the output of the third layer while $f_r(\X_{0,j*},\W^+)$ is the output of the sixth layer.

From the perspective of optimization, the third term in the objective function (\ref{eq:L}) above is equivalent to a multi-layer perceptron using the latent item vectors $\v_j$ as target while the fourth term is equivalent to an SDAE minimizing the reconstruction error. Seeing from the view of neural networks (NN), when $\lambda_s$ approaches positive infinity, training of the probabilistic graphical model of CDL in Figure \ref{fig:cdl_pgm}(left) would degenerate to simultaneously training two neural networks overlaid together with a common input layer (the corrupted input) but different output layers, as shown in Figure \ref{fig:twonet}. Note that the second network is much more complex than typical neural networks due to the involvement of the rating matrix.

When the ratio $\lambda_n/\lambda_v$ approaches positive infinity, it will degenerate to a two-step model in which the latent representation learned using SDAE is put directly into the CTR.  Another extreme happens when $\lambda_n/\lambda_v$ goes to zero where the decoder of the SDAE essentially vanishes.  On the right of Figure \ref{fig:cdl_pgm} is the graphical model of the degenerated CDL when $\lambda_n/\lambda_v$ goes to zero.
As demonstrated in the experiments, the predictive performance will suffer greatly for both extreme cases.

For $\u_i$ and $\v_j$, coordinate ascent similar to \cite{DBLP:conf/kdd/WangB11,DBLP:conf/icdm/HuKV08} is used. Given the current $\W^+$, we compute the gradients of $\mathscr{L}$ with respect to $\u_i$ and $\v_j$ and set them to zero, leading to the following update rules:
\begin{align}
\u_i&\leftarrow(\V \C_i \V^T+\lambda_u \I_K)^{-1}\V \C_i \R_i \nonumber \\
\v_j&\leftarrow(\U \C_i \U^T+\lambda_v \I_K)^{-1}(\U \C_j \R_j+\lambda_v f_e(\X_{0,j*},\W^+)^T), \nonumber
\end{align}
where $\U=(\u_i)^I_{i=1}$, $\V=(\v_j)^J_{j=1}$, $\C_i = \mbox{diag}(\C_{i1},\ldots,\C_{iJ})$ is a diagonal matrix,
$\R_i = (\R_{i1},\ldots,\R_{iJ})^T$ is a column vector containing all the ratings of user $i$,
and $\C_{ij}$ reflects the confidence controlled by $a$ and $b$ as discussed in \cite{DBLP:conf/icdm/HuKV08}.

Given $\U$ and $\V$, we can learn the weights $\W_l$ and biases $\b_l$ for each layer using the back-propagation learning algorithm. The gradients of the likelihood with respect to $\W_l$ and $\b_l$ are as follows:
\begin{align*}
&\nabla_{\W_l}\mathscr{L} = -\lambda_w\W_l \\
&-\lambda_v\sum\limits_j\nabla_{\W_l}f_e(\X_{0,j*},\W^+)^T(f_e(\X_{0,j*},\W^+)^T-\v_j)\\
&-\lambda_n\sum\limits_j\nabla_{\W_l}f_r(\X_{0,j*},\W^+)(f_r(\X_{0,j*},\W^+)-\X_{c,j*})
\end{align*}
\begin{align*}
&\nabla_{\b_l}\mathscr{L} = -\lambda_w\b_l \\
&-\lambda_v\sum\limits_j\nabla_{\b_l}f_e(\X_{0,j*},\W^+)^T(f_e(\X_{0,j*},\W^+)^T-\v_j)\\
&-\lambda_n\sum\limits_j\nabla_{\b_l}f_r(\X_{0,j*},\W^+)(f_r(\X_{0,j*},\W^+)-\X_{c,j*}).
\end{align*}
By alternating the update of $\U$, $\V$, $\W_l$, and $\b_l$, we can find a local optimum for $\mathscr{L}$. Several commonly used techniques such as using a momentum term may be used to alleviate the local optimum problem. For completeness, we also provide a sampling- based algorithm for CDL in the appendix.

\subsection{Prediction}

Let $D$ be the observed test data. Similar to \cite{DBLP:conf/kdd/WangB11}, we use the point estimates of $\u_i$, $\W^+$
and $\ep_j$ to calculate the predicted rating:
\begin{align}
E[\R_{ij}|D]\approx E[\u_i|D]^T(E[f_e(\X_{0,j*},\W^+)^T|D]+E[\ep_j|D]),\nonumber
\end{align}
where $E[\cdot]$ denotes the expectation operation.
In other words, we approximate the predicted rating as:
\begin{align}
\R^*_{ij}\approx(\u^*_j)^T(f_e(\X_{0,j*},{\W^+}^*)^T+\ep^*_j)=(\u^*_i)^T\v^*_j .\nonumber
\end{align}
Note that for any new item $j$ with no rating in the training data, its offset $\ep^*_j$ will be $\0$.

\section{Experiments}
\begin{figure*}[!tb]
\begin{center}
\subfigure{
\includegraphics[height=4.0cm]{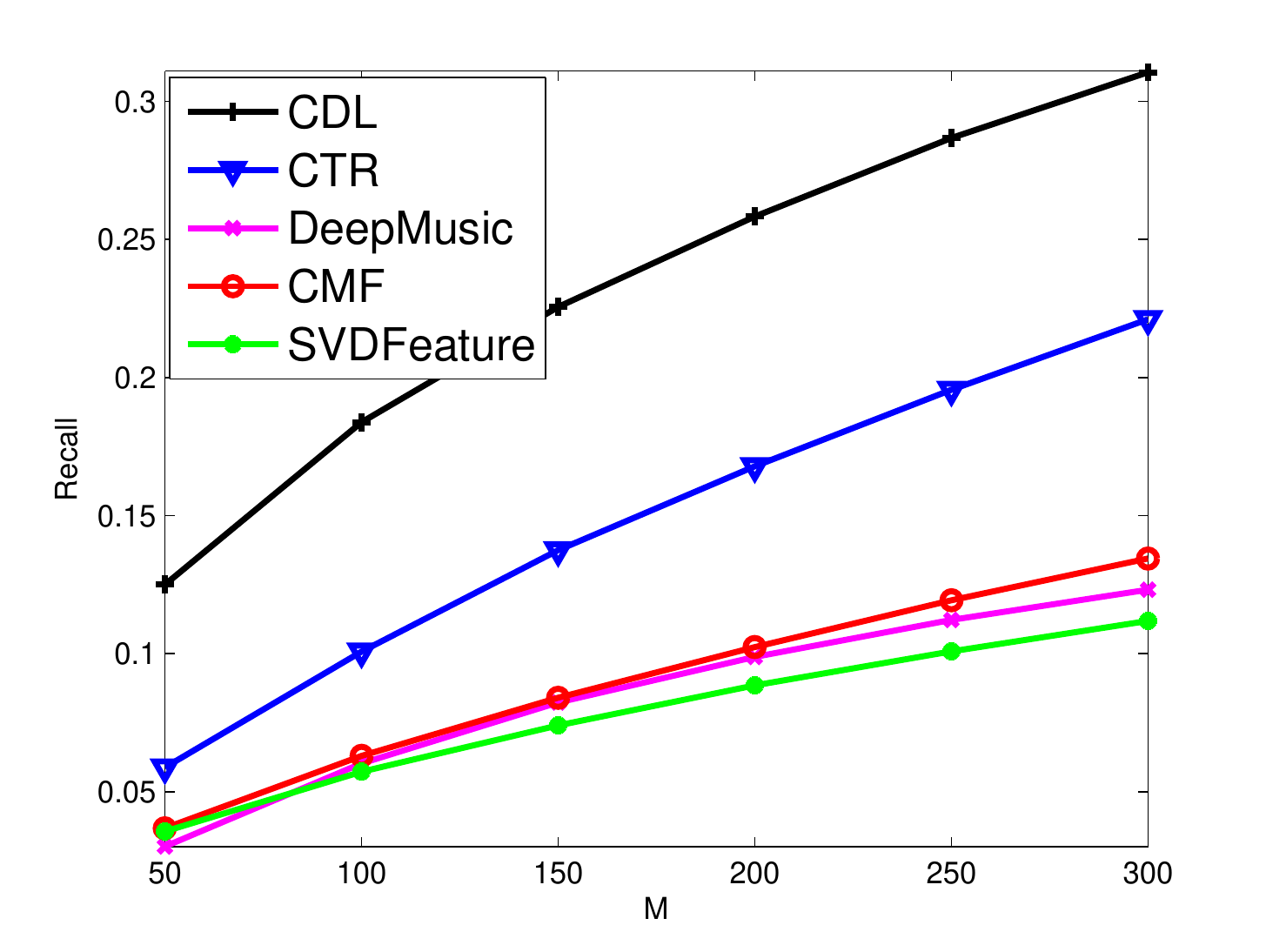}}
\hspace{0.05in}
\subfigure{
\includegraphics[height=4.0cm]{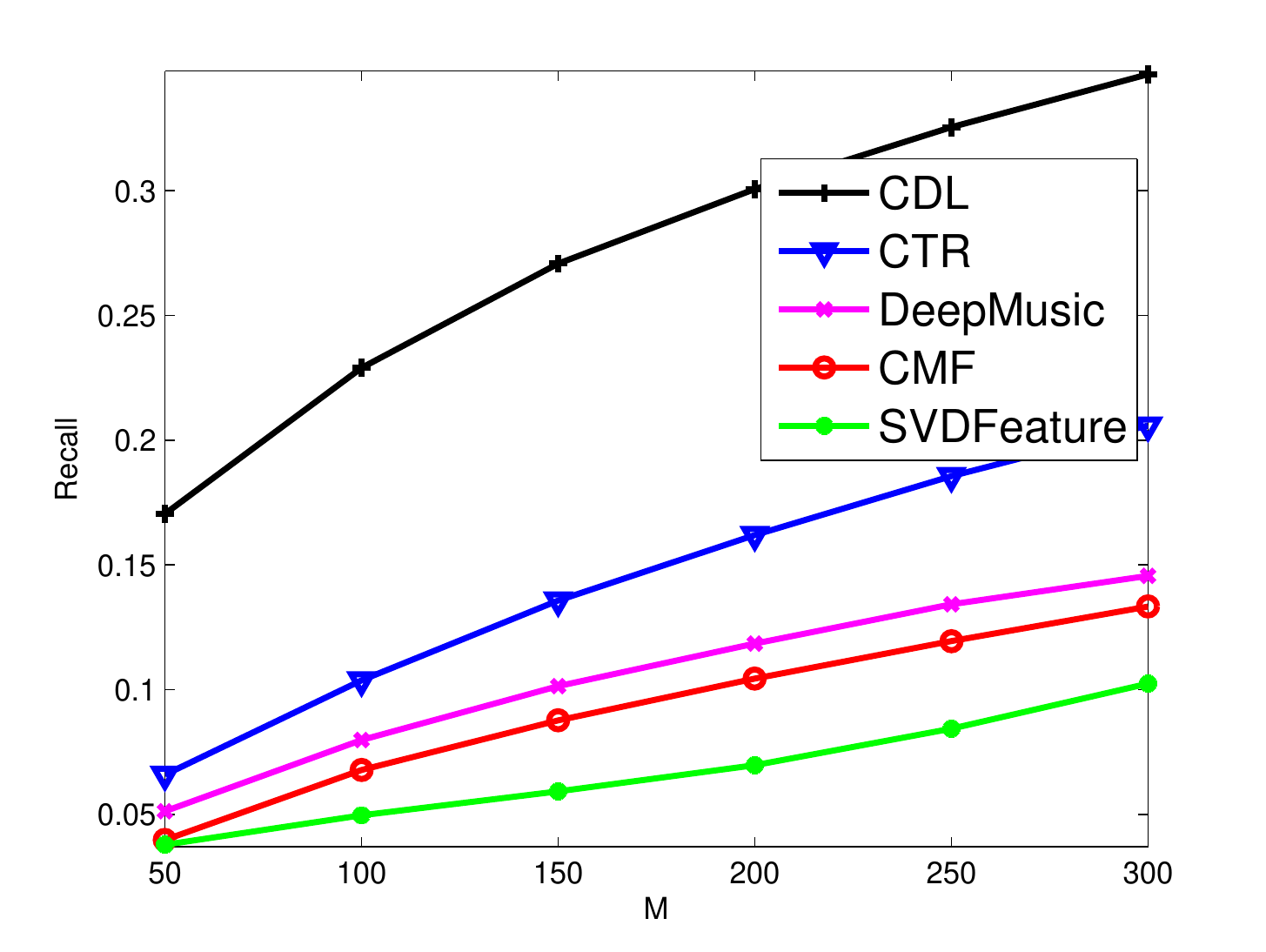}}
\hspace{0.05in}
\subfigure{
\includegraphics[height=4.0cm]{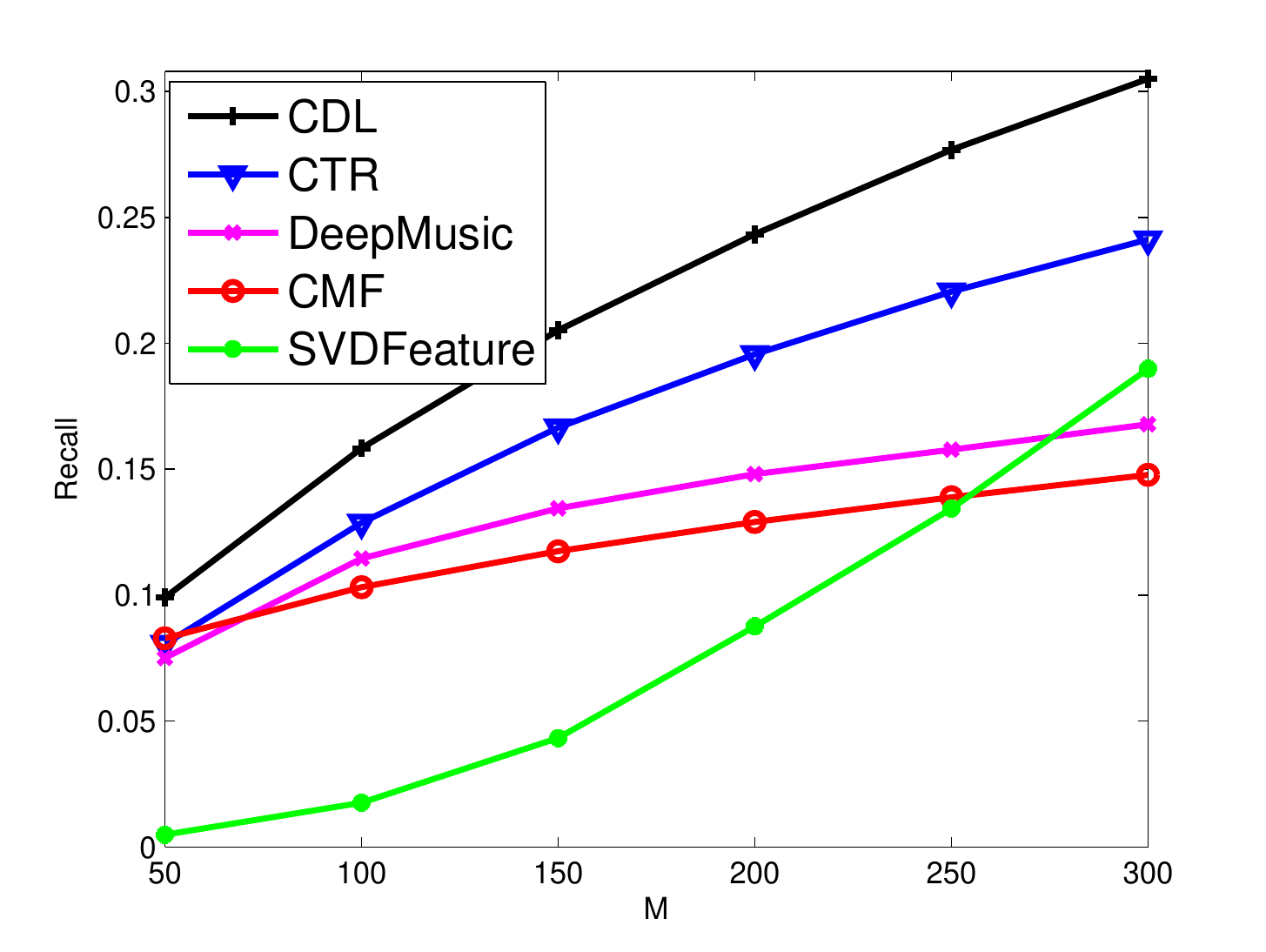}}
\end{center}
\vskip -0.25in
\caption{Performance comparison of CDL, CTR, DeepMusic, CMF, and SVDFeature based on recall@$M$ for datasets \emph{citeulike-a}, \emph{citeulike-t}, and \emph{Netflix} in the sparse setting. A 2-layer CDL is used.}
\vskip -0.0in
\label{fig:sparse}
\vskip -0.1in
\end{figure*}

\begin{figure*}[!tb]
\begin{center}
\subfigure{
\includegraphics[height=4.0cm]{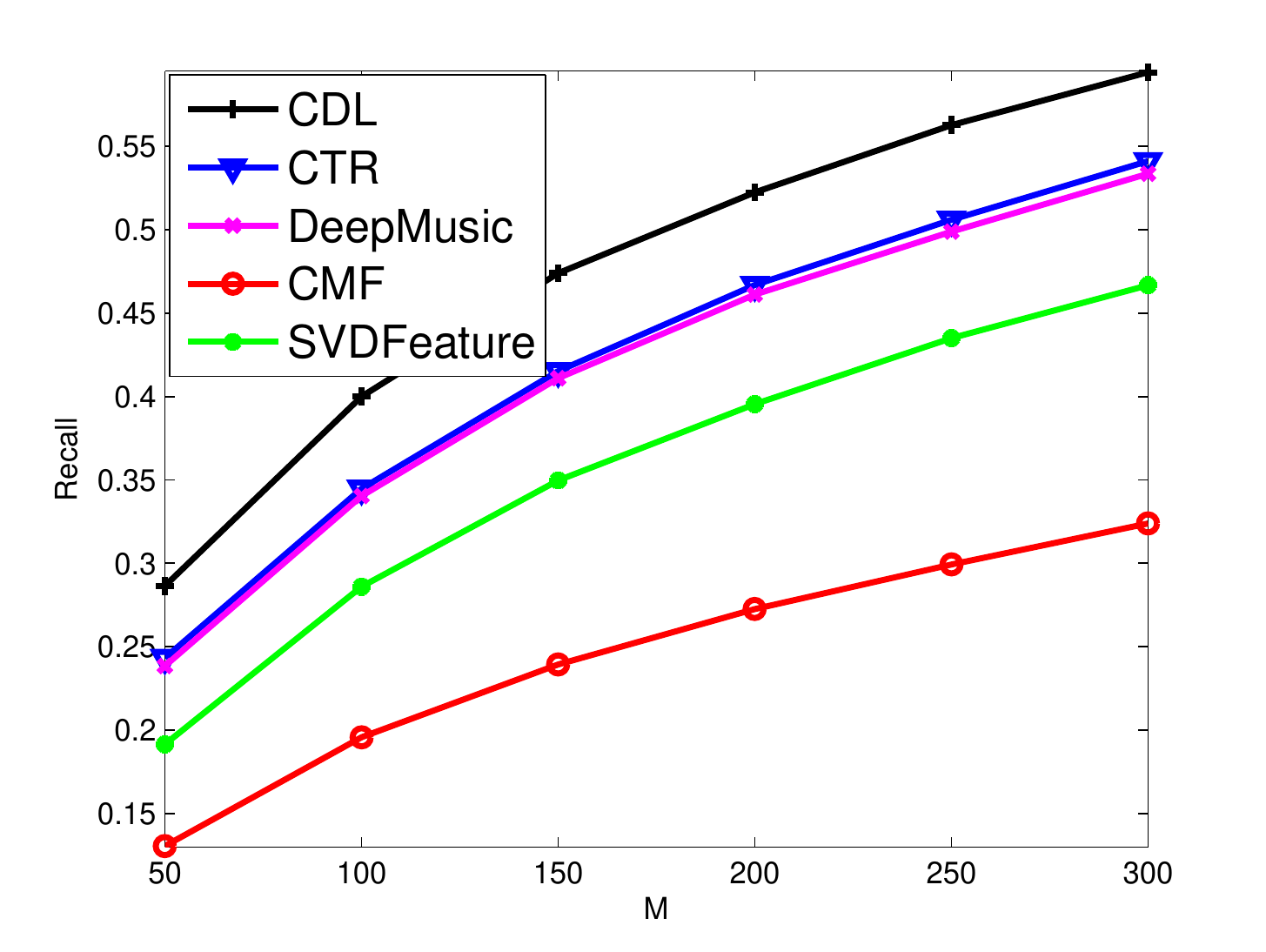}}
\hspace{0.05in}
\subfigure{
\includegraphics[height=4.0cm]{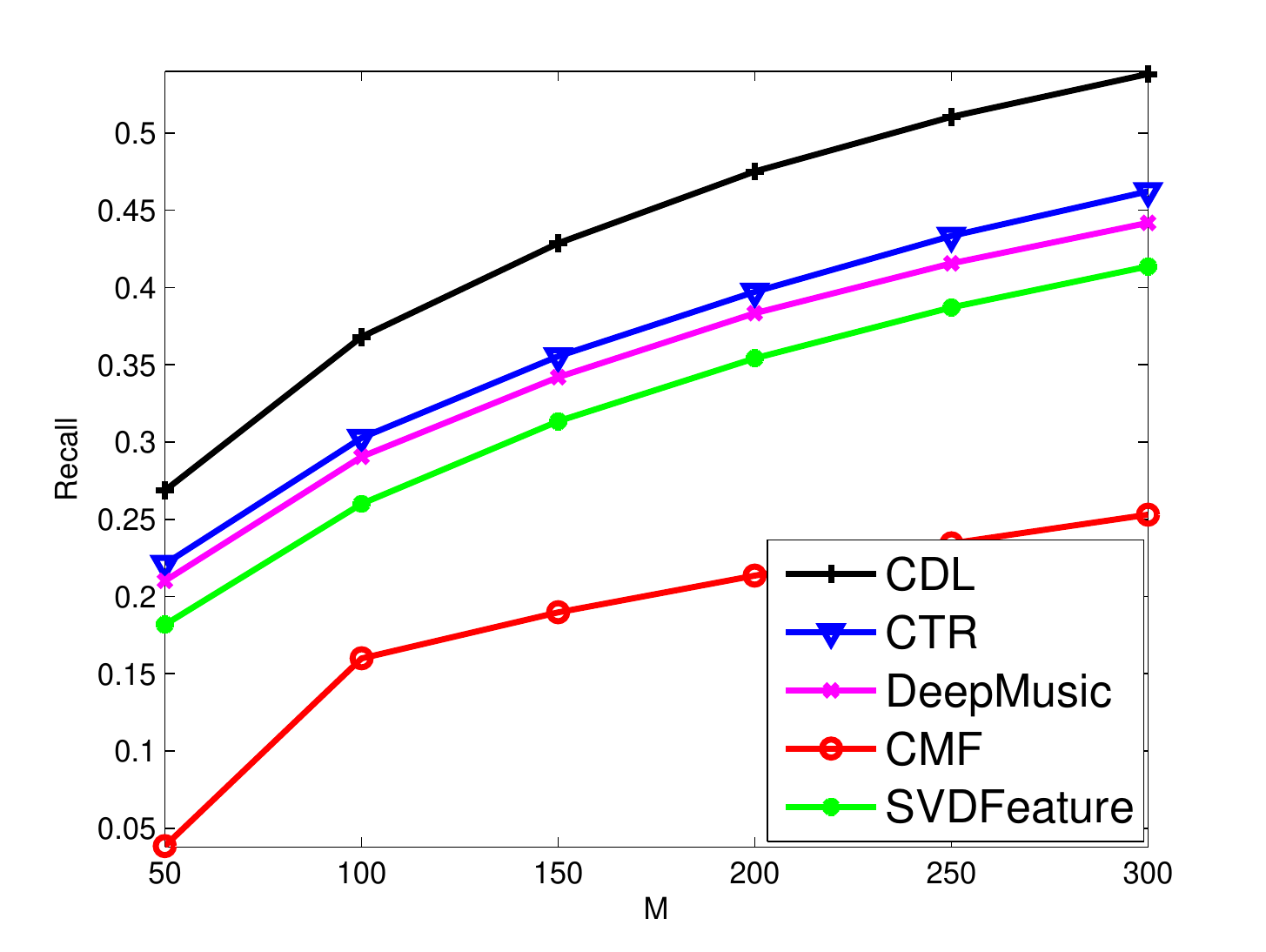}}
\hspace{0.05in}
\subfigure{
\includegraphics[height=4.0cm]{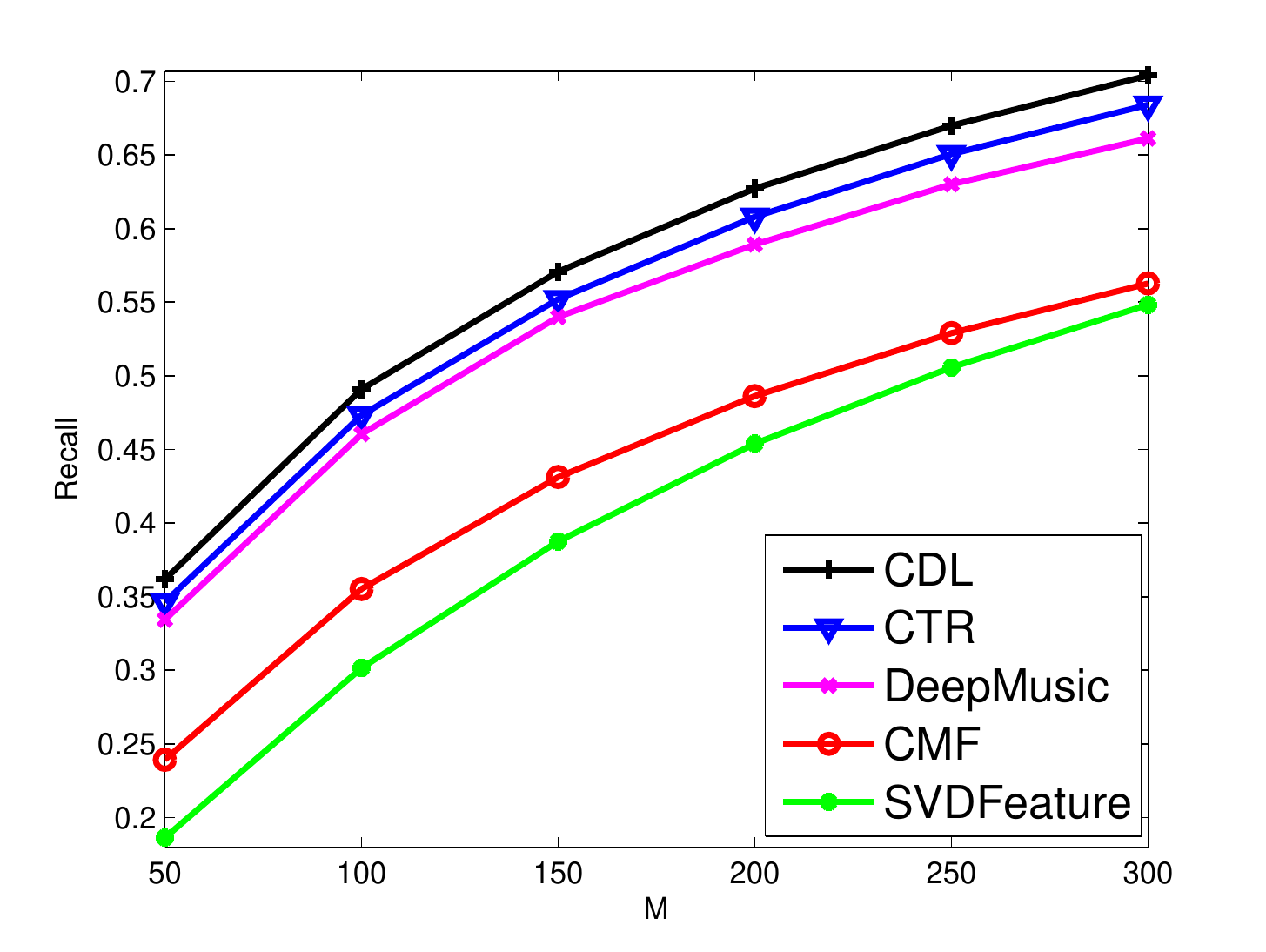}}
\end{center}
\vskip -0.25in
\caption{Performance comparison of CDL, CTR, DeepMusic, CMF, and SVDFeature based on recall@$M$ for datasets \emph{citeulike-a}, \emph{citeulike-t}, and \emph{Netflix} in the dense setting. A 2-layer CDL is used.}
\vskip -0.0in
\label{fig:dense}
\vskip -0.2in
\end{figure*}

Extensive experiments are conducted on three real-world datasets from different domains to demonstrate the effectiveness of our model both quantitatively and qualitatively\footnote{Code and data are available at \url{www.wanghao.in}}.

\subsection{Datasets}

We use three datasets from different real-world domains, two from CiteULike\footnote{CiteULike allows users to create their own collections of articles. There are abstract, title, and tags for each article. More details about the CiteULike data can be found at \url{http://www.citeulike.org}.} and one from Netflix, for our experiments.  The first two datasets, from \cite{DBLP:conf/ijcai/WangCL13}, were collected in different ways, specifically, with different scales and different degrees of sparsity to mimic different practical situations.  The first dataset, \emph{citeulike-a}, is mostly from \cite{DBLP:conf/kdd/WangB11}.  The second dataset, \emph{citeulike-t}, was collected independently of the first one.  They manually selected $273$ seed tags and collected all the articles with at least one of those tags. Similar to \cite{DBLP:conf/kdd/WangB11}, users with fewer than $3$ articles are not included. As a result, \emph{citeulike-a} contains $5551$ users and $16980$ items. For \mbox{\emph{citeulike-t}}, the numbers are $7947$ and $25975$.
We can see that \emph{citeulike-t} contains more users and items than \mbox{\emph{citeulike-a}}.  Also, \emph{citeulike-t} is much sparser as only $0.07\%$ of its user-item matrix entries contain ratings but \mbox{\emph{citeulike-a}} has ratings in $0.22\%$ of its user-item matrix entries.

The last dataset, \emph{Netflix}, consists of two parts. The first part, with ratings and movie titles, is from the Netflix challenge dataset. The second part, with plots of the corresponding movies, was collected by us from IMDB \footnote{\url{http://www.imdb.com}}. Similar to \cite{DBLP:conf/kdd/ZhouZ12}, in order to be consistent with the implicit feedback setting of the first two datasets, we extract only positive ratings (rating $5$) for training and testing. After removing users with less than $3$ positive ratings and movies without plots, we have $407261$ users, $9228$ movies, and $15348808$ ratings in the final dataset.

We follow the same procedure as that in \cite{DBLP:conf/kdd/WangB11} to preprocess the text information (item content) extracted from the titles and abstracts of the articles and the plots of the movies.  After removing stop words, the top $S$ discriminative words according to the tf-idf values are chosen to form the vocabulary ($S$ is $8000$, $20000$, and $20000$ for the three datasets).

\subsection{Evaluation Scheme}

For each dataset, similar to \cite{DBLP:conf/ijcai/WangCL13,DBLP:journals/tkde/WangL15}, we randomly select $P$ items associated with each user to form the training set and use all the rest of the dataset as the test set. To evaluate and compare the models under both sparse and dense settings, we set $P$ to $1$ and $10$, respectively, in our experiments. For each value of $P$, we repeat the evaluation five times with different randomly selected training sets and the average performance is reported.

As in \cite{DBLP:conf/kdd/WangB11,DBLP:conf/icml/PurushothamL12,DBLP:conf/ijcai/WangCL13}, we use recall as the performance measure because the rating information is in the form of implicit feedback \cite{DBLP:conf/icdm/HuKV08,DBLP:conf/uai/RendleFGS09}.  Specifically, a zero entry may be due to the fact that the user is not interested in the item, or that the user is not aware of its existence.  As such, precision is not a suitable performance measure.  Like most recommender systems, we sort the predicted ratings of the candidate items and recommend the top $M$ items to the target user. The recall@$M$ for each user is then defined as:
\begin{align}
\mbox{recall@$M$}=\frac{\scriptsize \mbox{number of items that the user likes among the top $M$}}{\scriptsize \mbox{total number of items that the user likes}} . \nonumber
\end{align}
The final result reported is the average recall over all users.

Another evaluation metric is the mean average precision (mAP). Exactly the same as \cite{DBLP:conf/nips/OordDS13}, we set the cutoff point at $500$ for each user.

\subsection{Baselines and Experimental Settings}

The models included in our comparison are listed as follows:
\begin{compactitem}
\item \textbf{CMF}: Collective Matrix Factorization \cite{DBLP:conf/kdd/SinghG08} is a model incorporating different sources of information by simultaneously factorizing multiple matrices. In this paper, the two factorized matrices are $\R$ and $\X_c$.
\item \textbf{SVDFeature}: SVDFeature \cite{DBLP:journals/jmlr/ChenZLCZY12} is a model for feature-based collaborative filtering. In this paper we use the content information $\X_c$ as raw features to feed into SVDFeature.
\item \textbf{DeepMusic}: DeepMusic \cite{DBLP:conf/nips/OordDS13} is a model for music recommendation mentioned in Section \ref{sec:intro}. We use the variant, a loosely coupled method, that achieves the best performance as our baseline.
\item \textbf{CTR}: Collaborative Topic Regression \cite{DBLP:conf/kdd/WangB11} is a model performing topic modeling and collaborative filtering simultaneously as mentioned in the previous section.
\item \textbf{CDL}: Collaborative Deep Learning is our proposed model as described above.  It allows different levels of model complexity by varying the number of layers.
\end{compactitem}

In the experiments, we first use a validation set to find the optimal hyperparameters for CMF, SVDFeature, CTR, and DeepMusic. For CMF, we set the regularization hyperparameters for the latent factors of different contexts to $10$. After the grid search, we find that CMF performs best when the weights for the rating matrix and content matrix (BOW) are both $5$ in the sparse setting. For the dense setting the weights are $8$ and $2$, respectively. For SVDFeature, the best performance is achieved when the regularization hyperparameters for the users and items are both $0.004$ with the learning rate equal to $0.005$. For DeepMusic, we find that the best performance is achieved using a CNN with two convolutional layers. We also try our best to tune the other hyperparameters. For CTR, we find that it can achieve good prediction performance when $\lambda_u=0.1$, $\lambda_v=10$, $a=1$, $b=0.01$, and $K=50$ (note that $a$ and $b$ determine the confidence parameters $\C_{ij}$). For CDL, we directly set $a=1$, $b=0.01$, $K=50$ and perform grid search on the hyperparameters $\lambda_u$, $\lambda_v$, $\lambda_n$, and $\lambda_w$.
For the grid search, we split the training data and use 5-fold cross validation.

\begin{table}[tb]
\centering
\caption{mAP for three datasets }\label{table:mapsparse}
\vskip 0.2cm
\begin{tabular}{|l|c|c|c|} \hline
  & \emph{citeulike-a} & \emph{citeulike-t}& \emph{Netflix}\\ \hline
CDL &\textbf{0.0514}	&\textbf{0.0453}&	\textbf{0.0312}\\ \hline
CTR & 0.0236 &	0.0175	&0.0223\\ \hline
DeepMusic & 0.0159 &	0.0118	&0.0167 \\ \hline
CMF & 0.0164 &	0.0104	&0.0158 \\ \hline
SVDFeature & 0.0152 &	0.0103	&0.0187 \\ \hline
\end{tabular}
\vskip -0.4cm
\end{table}

We use a masking noise
with a noise level of $0.3$ to get the corrupted input $\X_0$ from the clean input $\X_c$.  For CDL with more than one layer of SDAE ($L>2$), we use a dropout rate \cite{DBLP:conf/nips/BaldiS13,DBLP:conf/nips/WagerWL13,DBLP:journals/corr/abs-1207-0580} of $0.1$ to achieve adaptive regularization. In terms of network architecture, the number of hidden units $K_l$ is set to $200$ for $l$ such that $l \ne L/2$ and $0<l<L$.
While both $K_0$ and $K_L$ are equal to the number of words $S$ in the dictionary, $K_{L/2}$ is set to $K$ which is the number of dimensions of the learned representation. For example, the 2-layer CDL model ($L=4$) has a Bayesian SDAE of architecture `8000-200-50-200-8000' for the \emph{citeulike-a} dataset.

\subsection{Quantitative Comparison}

Figures \ref{fig:sparse} and \ref{fig:dense} show the results that compare CDL, CTR, DeepMusic, CMF, and SVDFeature using the three datasets under both the sparse ($P=1$) and dense ($P=10$) settings.  We can see that CTR is a strong baseline which beats DeepMusic, CMF, and SVDFeature in all datasets even though DeepMusic has a deep architecture. In the sparse setting, CMF outperforms SVDFeature most of the time and sometimes even achieves performance comparable to CTR. DeepMusic performs poorly due to lack of ratings and overfitting. In the dense setting, SVDFeature is significantly better than CMF for \emph{citeulike-a} and \emph{citeulike-t} but is inferior to CMF for \emph{Netflix}. DeepMusic is still slightly worse than CTR due to the reasons mentioned in \mbox{Section \ref{sec:intro}}. To focus more specifically on comparing CDL with CTR, we can see that for \emph{citeulike-a}, \mbox{2-layer} CDL outperforms CTR by a margin of 4.2\%$\sim$6.0\% in the sparse setting and 3.3\%$\sim$4.6\% in the dense setting.  If we increase the number of layers to $3$ ($L=6$), the margin will go up to 5.8\%$\sim$8.0\% and 4.3\%$\sim$5.8\%, respectively. Similarly for \emph{citeulike-t}, \mbox{2-layer CDL} outperforms CTR by a margin of 10.4\%$\sim$13.1\% in the sparse setting and 4.7\%$\sim$7.6\% in the dense setting. When the number of layers is increased to $3$, the margin will even go up to 11.0\%$\sim$14.9\% and 5.2\%$\sim$8.2\%, respectively.  For \emph{Netflix}, 2-layer CDL outperforms CTR by a margin of 1.9\%$\sim$5.9\% in the sparse setting and 1.5\%$\sim$2.0\% in the dense setting. As we can see, seamless and successful integration of deep learning and RS requires careful designs to avoid overfitting and achieve significant performance boost.

Table \ref{table:mapsparse} shows the mAP for all models in the sparse settings. We can see that the mAP of CDL is almost or more than twice of CTR. Tables \ref{table:sparse} and \ref{table:dense} show the recall@300 results when CDL with different numbers of layers are applied to the three datasets under both the sparse and dense settings.  As we can see, for \emph{\mbox{citeulike-t}} and \emph{Netflix}, the recall increases as the number of layers increases.  For \mbox{\emph{citeulike-a}}, CDL starts to overfit when it exceeds two layers.  Since the standard deviation is always very small ($4.31\times10^{-5} \sim 9.31\times10^{-3}$), we do not include it in the figures and tables as it is not noticeable anyway.

\begin{table}[tb]
\centering
\caption{Recall@300 in the sparse setting (\%)}\label{table:sparse}
\vskip 0.2cm
\begin{tabular}{|l|c|c|c|} \hline
\#layers &1&2&3\\ \hline
\emph{citeulike-a} & 27.89 & \textbf{31.06} & 30.70 \\ \hline
\emph{citeulike-t} & 32.58 & 34.67 & \textbf{35.48} \\ \hline
\emph{Netflix} & 29.20 &  30.50 & \textbf{31.01} \\ \hline
\end{tabular}
\vskip -0.5cm
\end{table}

Note that the results are somewhat different for the first two datasets although they are from the same domain.  This is due to the different ways in which the datasets were collected, as discussed above.  Specifically, both the text information
and the rating matrix in \emph{citeulike-t} are much sparser.\footnote{Each article in \emph{citeulike-a} has $66.6$ words on average and that for \emph{citeulike-t} is $18.8$.}
By seamlessly integrating deep representation learning for content information and CF for the rating matrix, CDL can handle both the sparse rating matrix and the sparse text information much better and learn a much more effective latent representation for each item and hence each user.
\begin{figure}[!tb]
\begin{center}
\subfigure{
\includegraphics[height=3.1cm]{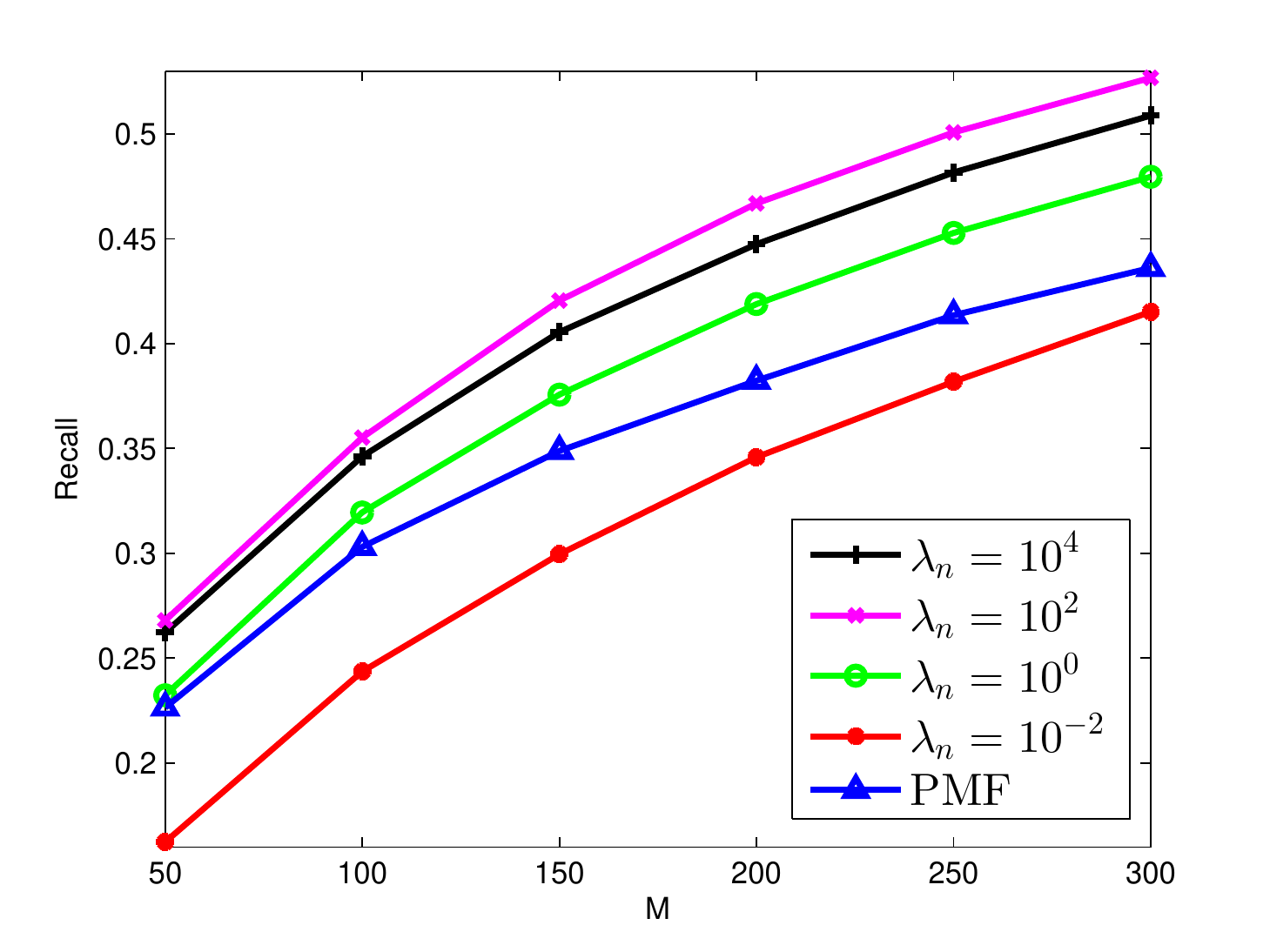}}
\hspace{-0.1in}
\subfigure{
\includegraphics[height=3.1cm]{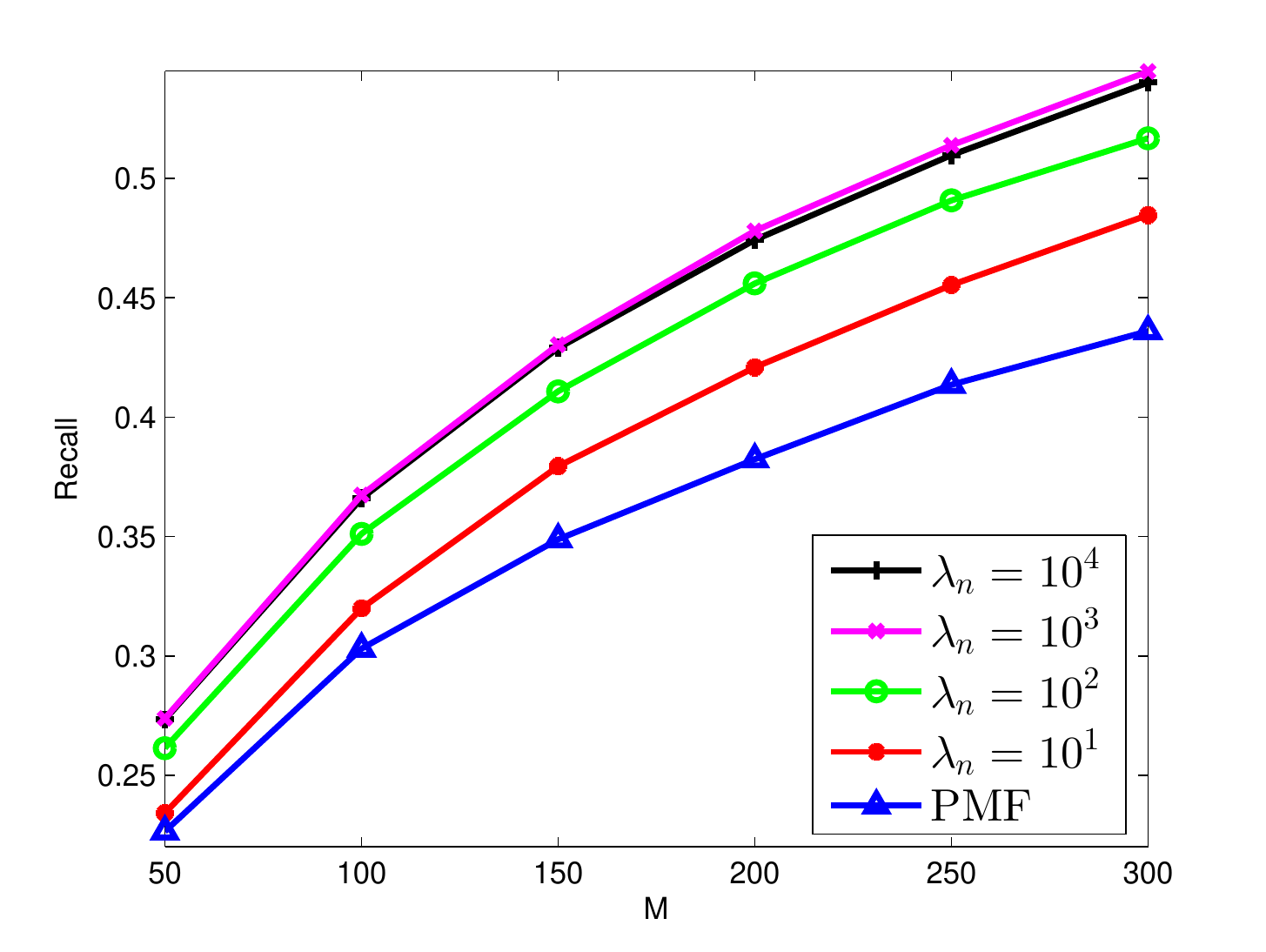}}
\end{center}
\vskip -0.3in
\caption{Performance of CDL based on recall@$M$ for different values of $\lambda_n$ on \emph{citeulike-t}. The left plot is for $L=2$ and the right one is for $L=6$.}
\label{fig:param}
\vskip -0.2in
\end{figure}

Figure \ref{fig:param} shows the results for different values of $\lambda_n$ using \emph{\mbox{citeulike-t}} under the dense setting.  We set $\lambda_u=0.01$, $\lambda_v=100$, and $L$ to $2$ and $6$.  Similar phenomena are observed when the number of layers and the value of $P$ are varied but they are omitted here due to space constraints.  As mentioned in the previous section, when $\lambda_n$ is extremely large, $\lambda_n/\lambda_v$ will approach positive infinity so that CDL degenerates to two separate models.  In this case the latent item representation will be learned by the SDAE in an unsupervised manner and then it will be put directly into (a simplified version of) the CTR.  Consequently, there is no interaction between the Bayesian SDAE and the collaborative filtering component based on matrix factorization and hence the prediction performance will suffer greatly.  For the other extreme when $\lambda_n$ is extremely small, $\lambda_n/\lambda_v$ will approach zero so that CDL degenerates to that in Figure \ref{fig:cdl_pgm} in which the decoder of the Bayesian SDAE component essentially vanishes. This way the encoder of the Bayesian SDAE component will easily overfit the latent item vectors learned by simple matrix factorization. As we can see in Figure \ref{fig:param}, the prediction performance degrades significantly as $\lambda_n$ gets very large or very small. When $\lambda_n<0.1$, the recall@$M$ is already very close to (or even worse than) the result of PMF.

\subsection{Qualitative Comparison}

To gain a better insight into CDL, we first take a look at two example users in the \emph{citeulike-t} dataset and represent the profile of each of them using the top three matched topics.  We examine the top 10 recommended articles returned by a 3-layer~($L=6$) CDL and CTR.  The models are trained under the sparse setting ($P=1$).  From Table \ref{table:userprof}, we can speculate that user I might be a computer scientist with focus on tag recommendation, as clearly indicated by the first topic in CDL and the second one in CTR. CDL correctly recommends many articles on tagging systems while CTR focuses on social networks instead.  When digging into the data, we find that the only rated article in the training data is `What drives content tagging: the case of photos on Flickr', which is an article that talks about the impact of social networks on tagging behaviors.  This may explain why CTR focuses its recommendation on social networks. On the other hand, CDL can better understand the key points of the article (i.e., tagging and CF) to make appropriate recommendation accordingly. Consequently, the precision of CDL and CTR is 70\% and 10\%, respectively.

\begin{table}[tb]
\centering
\caption{Recall@300 in the dense setting (\%)}\label{table:dense}
\vskip 0.2cm
\begin{tabular}{|l|c|c|c|} \hline
\#layers &1&2&3\\ \hline
\emph{citeulike-a} & 58.35 & \textbf{59.43} & 59.31 \\ \hline
\emph{citeulike-t} & 52.68 & 53.81 & \textbf{54.48} \\ \hline
\emph{Netflix} & 69.26 &  70.40 & \textbf{70.42} \\ \hline
\end{tabular}
\vskip -0.5cm
\end{table}

\begin{table*}[!t]
\caption{\small Interpretability of the latent structures learned
}\label{table:userprof}
\begin{center}
\begin{scriptsize}
\begin{tabular}{l|l|c}
\hline
  & user I (CDL) & in user's lib?\\
\hline
\multirow{3}{*}{top 3 topics}
& 1. search, image, query, images, queries, tagging, index, tags, searching, tag   &  \\
& 2. social, online, internet, communities, sharing, networking, facebook, friends, ties, participation  &\\
& 3. collaborative, optimization, filtering, recommendation, contextual, planning, items, preferences  &\\
\hline
\multirow{10}{*}{top 10 articles}
& \textbf{1. The structure of collaborative tagging Systems} & \textbf{yes} \\
& \textbf{2. Usage patterns of collaborative tagging systems} & \textbf{yes} \\
& 3. Folksonomy as a complex network & no\\
& \textbf{4. HT06, tagging paper, taxonomy, Flickr, academic article, to read} & \textbf{yes} \\
& \textbf{5. Why do tagging systems work} & \textbf{yes} \\
& 6. Information retrieval in folksonomies: search and ranking & no \\
& \textbf{7. tagging, communities, vocabulary, evolution} & \textbf{yes} \\
& \textbf{8. The complex dynamics of collaborative tagging} & \textbf{yes} \\
& 9. Improved annotation of the blogosphere via autotagging and hierarchical clustering & no \\
& \textbf{10. Collaborative tagging as a tripartite network} & \textbf{yes} \\
\hline \hline
 & user I (CTR) & in user's lib?\\
\hline
\multirow{3}{*}{top 3 topics}
& 1. social, online, internet, communities, sharing, networking, facebook, friends, ties, participation&\\
& 2. search, image, query, images, queries, tagging, index, tags, searching, tag&\\
& 3. feedback, event, transformation, wikipedia, indicators, vitamin, log, indirect, taxonomy &\\
\hline
\multirow{10}{*}{top 10 articles}
& \textbf{1. HT06, tagging paper, taxonomy, Flickr, academic article, to read} & \textbf{yes}\\
& 2. Structure and evolution of online social networks & no\\
& 3. Group formation in large social networks: membership, growth, and evolution & no\\
& 4. Measurement and analysis of online social networks & no\\
& 5. A face(book) in the crowd: social searching vs. social browsing & no\\
& 6. The strength of weak ties & no\\
& 7. Flickr tag recommendation based on collective knowledge & no\\
& 8. The computer-mediated communication network & no\\
& 9. Social capital, self-esteem, and use of online social network sites: A longitudinal analysis &no\\
& 10. Increasing participation in online communities: A framework for human-computer interaction &no\\
\hline \hline
 & user II (CDL) & in user's lib?\\
\hline
\multirow{3}{*}{top 3 topics}
& 1. flow, cloud, codes, matter, boundary, lattice, particles, galaxies, fluid, galaxy & \\
& 2. mobile, membrane, wireless, sensor, mobility, lipid, traffic, infrastructure, monitoring, ad & \\
& 3. hybrid, orientation, stress, fluctuations, load, temperature, centrality, mechanical, two-dimensional, heat & \\
\hline
\multirow{10}{*}{top 10 articles}
& \textbf{1. Modeling the flow of dense suspensions of deformable particles in three dimensions} & \textbf{yes}\\
& \textbf{2. Simplified particulate model for coarse-grained hemodynamics simulations} & \textbf{yes}\\
& \textbf{3. Lattice Boltzmann simulations of blood flow: non-newtonian rheology and clotting processes} & \textbf{yes}\\
& \textbf{4. A genome-wide association study for celiac disease identifies risk variants} & \textbf{yes}\\
& \textbf{5. Efficient and accurate simulations of deformable particles} &\textbf{yes}\\
& \textbf{6. A multiscale model of thrombus development} & \textbf{yes}\\
& \textbf{7. Multiphase hemodynamic simulation of pulsatile flow in a coronary artery} & \textbf{yes}\\
& \textbf{8. Lattice Boltzmann modeling of thrombosis in giant aneurysms} & \textbf{yes}\\
& \textbf{9. A lattice Boltzmann simulation of clotting in stented aneursysms} & \textbf{yes}\\
& \textbf{10. Predicting dynamics and rheology of blood flow} & \textbf{yes}\\
\hline \hline
 & user II (CTR) & in user's lib?\\
\hline
\multirow{3}{*}{top 3 topics}
& 1. flow, cloud, codes, matter, boundary, lattice, particles, galaxies, fluid, galaxy & \\
& 2. transition, equations, dynamical, discrete, equation, dimensions, chaos, transitions, living, trust & \\
& 3. mobile, membrane, wireless, sensor, mobility, lipid, traffic, infrastructure, monitoring, ad & \\
\hline
\multirow{10}{*}{top 10 articles}
& \textbf{1. Multiphase hemodynamic simulation of pulsatile flow in a coronary artery} & \textbf{yes}\\
& 2. The metallicity evolution of star-forming galaxies from redshift 0 to 3 & no\\
& 3. Formation versus destruction: the evolution of the star cluster population in galaxy mergers & no\\
& 4. Clearing the gas from globular clusters & no\\
& 5. Macroscopic effects of the spectral structure in turbulent flows & no\\
& 6. The WiggleZ dark energy survey & no\\
& 7. Lattice-Boltzmann simulation of blood flow in digitized vessel networks & no\\
& 8. Global properties of 'ordinary' early-type galaxies & no\\
& \textbf{9. Proteus : a direct forcing method in the simulations of particulate flows} & \textbf{yes}\\
& \textbf{10. Analysis of mechanisms for platelet near-wall excess under arterial blood flow conditions} & \textbf{yes}\\
\hline
\end{tabular}
\end{scriptsize}
\end{center}
\vskip -0.3in
\end{table*}

\begin{table*}[!thb]
\newcommand{\tabincell}[2]{\begin{tabular}{@{}#1@{}}#2\end{tabular}}
\caption{\small Example user with recommended movies
}\label{table:movierec} \vskip -0.0cm
\begin{center}
\begin{scriptsize}
\begin{tabular}{c|l|l|l}
\hline
\multirow{2}{*}{User III} & \multicolumn{3}{l}{Movies in the training set: \textbf{Moonstruck, True Romance, Johnny English, American Beauty, The}}\tabularnewline
 & \multicolumn{3}{l}{\textbf{Princess Bride, Top Gun, Double Platinum, Rising Sun, Dead Poets Society, Waiting for Guffman}}\tabularnewline
\hline
\hline
\# training samples & 2 & 4 & 10\tabularnewline
\hline
\multirow{10}{*}{\tabincell{l}{Top 10 recommended \\movies by CTR}} & Swordfish & \textbf{Pulp Fiction}  & \textbf{Best in Snow} \tabularnewline
\cline{2-4}
 & A Fish Called Wanda & A Clockwork Orange  & \textbf{Chocolat}\tabularnewline
\cline{2-4}
 & \textbf{Terminator 2}  & Being John Malkovich  & \textbf{Good Will Hunting} \tabularnewline
\cline{2-4}
 & A Clockwork Orange & \textbf{Raising Arizona}  & \textbf{Monty Python and the Holy Grail} \tabularnewline
\cline{2-4}
 & Sling Blade & Sling Blade  & Being John Malkovich \tabularnewline
\cline{2-4}
 & Bridget Jones's Diary & Swordfish  & \textbf{Raising Arizona}\tabularnewline
\cline{2-4}
 & \textbf{Raising Arizona} & A Fish Called Wanda  & The Graduate \tabularnewline
\cline{2-4}
 & A Streetcar Named Desire & Saving Grace  & Swordfish \tabularnewline
\cline{2-4}
 & The Untouchables & The Graduate  & Tootsie \tabularnewline
\cline{2-4}
 & The Full Monty & Monster's Ball  & Saving Private Ryan \tabularnewline

\hline
\hline
\# training samples & 2 & 4 & 10\tabularnewline
\hline
\multirow{10}{*}{\tabincell{l}{Top 10 recommended \\movies by CDL}} & Snatch & \textbf{Pulp Fiction} & \textbf{Good Will Hunting} \tabularnewline
\cline{2-4}
 & \textbf{The Big Lebowski} & Snatch  & \textbf{Best in Show}\tabularnewline
\cline{2-4}
 & \textbf{Pulp Fiction}  & \textbf{The Usual Suspect} & \textbf{The Big Lebowski} \tabularnewline
\cline{2-4}
 & Kill Bill & Kill Bill & \textbf{A Few Good Men} \tabularnewline
\cline{2-4}
 & \textbf{Raising Arizona} & Momento & \textbf{Monty Python and the Holy Grail}\tabularnewline
\cline{2-4}
 & The Big Chill & \textbf{The Big Lebowski} & \textbf{Pulp Fiction}\tabularnewline
\cline{2-4}
 & Tootsie & \textbf{One Flew Over the Cuckoo's Nest}  & The Matrix\tabularnewline
\cline{2-4}
 & Sense and Sensibility & As Good as It Gets & \textbf{Chocolat} \tabularnewline
\cline{2-4}
 & Sling Blade & \textbf{Goodfellas} & \textbf{The Usual Suspect}\tabularnewline
\cline{2-4}
 & Swinger & The Matrix  & \textbf{CaddyShack}\tabularnewline
\cline{2-4}

\hline
\end{tabular}
\end{scriptsize}
\end{center}
\vskip -0.2in
\end{table*}

From the matched topics returned by both CDL and CTR, user~II might be a researcher on blood flow dynamic theory particularly in the field of medical science.  CDL correctly captures the user profile and achieves a precision of 100\%.  However, CTR recommends quite a few articles on astronomy instead.  When examining the data, we find that the only rated article returned by CTR is `Simulating deformable particle suspensions using a coupled lattice-Boltzmann and finite-element method'.  As expected, this article is on deformable particle suspension and the flow of blood cells.  CTR might have misinterpreted this article, focusing its recommendation on words like `flows' and `formation' separately.  This explains why CTR recommends articles like `Formation versus destruction: the evolution of the star cluster population in galaxy mergers'~(formation) and `Macroscopic effects of the spectral structure in turbulent flows' (flows).  As a result, its precision is only 30\%.

From these two users, we can see that with a more effective representation, CDL can capture the key points of articles and the user preferences more accurately (e.g., user~I).  Besides, it can model the co-occurrence and relations of words better (e.g., user~II).

We next present another case study which is for the \emph{Netflix} dataset under the dense setting ($P=10$).  In this case study, we choose one user (user III) and vary the number of ratings (positive feedback) in the training set given by the user from $1$ to $10$. The partition of training and test data remains the same for all other users. This is to examine how the recommendation of CTR and CDL adapts as user III expresses preference for more and more movies. \mbox{Table \ref{table:movierec}} shows the recommendation lists of CTR and CDL when the number of training samples is set to $2$, $4$, and $10$.  When there are only two training samples, the two movies user III likes are `Moonstruck' and `True Romance', which are  both romance movies. For now the precision of CTR and CDL is close (20\% and 30\%).  When two more samples are added, the precision of CDL is boosted to 50\% while that of CTR remains unchanged (20\%).  That is because the two new movies, `Johnny English' and `American Beauty', belong to action and drama movies. CDL successfully captures the user's change of taste and gets two more recommendations right but CTR fails to do so.  Similar phenomena can be observed when the number of training samples increases from $4$ to $10$. From this case study, we can see that CDL is sensitive enough to changes of user taste and hence can provide more accurate recommendation.

\section{Complexity Analysis and Implementation}
Following the update rules in this paper, the computational complexity of updating $\u_i$ is $O(K^2J+K^3)$, where $K$ is the dimensionality of the learned representation and $J$ is the number of items. The complexity for $\v_j$ is $O(K^2I+K^3+SK_1)$, where $I$ is the number of users, $S$ is the size of the vocabulary, and $K_1$ is the dimensionality of the output in the first layer. Note that the third term $O(SK_1)$ is the cost of computing the output of the encoder and it is dominated by the computation of the first layer. For the update of all the weights and biases, the complexity is $O(JSK_1)$ since the computation is dominated by the first layer. Thus for a complete epoch the total time complexity is $O(JSK_1+K^2J^2+K^2I^2+K^3)$.

All our experiments are conducted on servers with $2$ Intel E5-2650 CPUs and $4$ NVIDIA Tesla M2090 GPUs each. Using the MATLAB implementation with GPU/C++ acceleration, each epoch takes only about $40$ seconds and each run takes $200$ epochs for the first two datasets. For \emph{Netflix} it takes about $60$ seconds per epoch and needs much fewer epochs (about $100$) to get satisfactory recommendation performance. Since \emph{Netflix} is much larger than the other two datasets, this shows that CDL is very scalable. We expect that changing the implementation to a pure C++/CUDA one would significantly reduce the time cost.

\section{Conclusion and Future Work}

We have demonstrated in this paper that state-of-the-art performance can be achieved by jointly performing deep representation learning for the content information and collaborative filtering for the ratings (feedback) matrix.  As far as we know, CDL is the first hierarchical Bayesian model to bridge the gap between state-of-the-art deep learning models and RS.  In terms of learning, besides the algorithm for attaining the MAP estimates, we also derive a sampling-based algorithm for the Bayesian treatment of CDL as a Bayesian generalized version of back-propagation.

Among the possible extensions that could be made to CDL, the bag-of-words representation may be replaced by more powerful alternatives, such as \cite{DBLP:conf/nips/MikolovSCCD13}.  The Bayesian nature of CDL also provides potential performance boost if other side information is incorporated as in \cite{DBLP:conf/aaai/WangSY15}.  Besides, as remarked above, CDL actually provides a framework that can also admit deep learning models other than SDAE.  One promising choice is the convolutional neural network model which, among other things, can explicitly take the context and order of words into account.  Further performance boost may be possible when using such deep learning models.

\section{Acknowledgments}
This research has been partially supported by research grant FSGRF14EG36.

%
\bibliographystyle{abbrv}
\bibliography{CDL}  
%
%
\appendix
\section{Bayesian Treatment for CDL}
For completeness we also derive a sampling-based algorithm for the Bayesian treatment of CDL.  It turns out to be a Bayesian and generalized version of the well-known back-propagation (BP) learning algorithm. Due to space constraints we only list the results here without detailed derivation.

\textbf{For $\W+$}:
We denote the concatenation of $\W_{l,*n}$ and $\b_l^{(n)}$ as $\W_{l,*n}^+$. Similarly, the concatenation of $\X_{l,j*}$ and $1$ is denoted as $\X_{l,j*}^+$. The subscripts of $\I$ are ignored. Then
\begin{align*}
&p(\W_{l,*n}^+|\X_{l-1,j*},\X_{l,j*},\lambda_s)\\
\propto \ & \NM(\W_{l,*n}^+|0,\lambda_w^{-1}\I) \NM(\X_{l,*n}|\sigma(\X_{l-1}^+\W_{l,*n}^+),\lambda_s^{-1}\I).
\end{align*}

\textbf{For $\X_{l,j*}$ ($l\neq L/2$)}:
Similarly, we denote the concatenation of $\W_{l}$ and $\b_l$ as $\W_{l}^+$ and have
\begin{align*}
&p(\X_{l,j*}|\W_l^+,\W_{l+1}^+,\X_{l-1,j*},\X_{l+1,j*}\lambda_s)\\
\propto \ & \NM(\X_{l,j*}|\sigma(\X_{l-1,j*}^+\W_l^+),\lambda_s^{-1}\I)\cdot\\
&\NM(\X_{l+1,j*}|\sigma(\X_{l,j*}^+\W_{l+1}^+),\lambda_s^{-1}\I).
\end{align*}
Note that for the last layer ($l=L$) the second Gaussian would be $\NM(\X_{c,j*}|\X_{l,j*},\lambda_s^{-1}\I)$ instead.

\textbf{For $\X_{l,j*}$ ($l= L/2$)}:
Similarly, we have
\begin{align*}
&p(\X_{l,j*}|\W_l^+,\W_{l+1}^+,\X_{l-1,j*},\X_{l+1,j*},\lambda_s,\lambda_v,\v_j)\\
\propto \ &\NM(\X_{l,j*}|\sigma(\X_{l-1,j*}^+\W_l^+),\lambda_s^{-1}\I)\cdot\\
&\NM(\X_{l+1,j*}|\sigma(\X_{l,j*}^+\W_{l+1}^+),\lambda_s^{-1}\I) \NM(\v_j|\X_{l,j*},\lambda_v^{-1}\I).
\end{align*}

\textbf{For $\v_j$}: The posterior $p(\v_j|\X_{L/2,j*},\R_{*j},\C_{*j},\lambda_v,\U)$
\begin{align*}
\propto\NM(\v_j|\X_{L/2,j*}^T,\lambda_v^{-1}\I)\prod\limits_i \NM(\R_{ij}|\u_i^T\v_j,\C_{ij}^{-1}).
\end{align*}

\textbf{For $\u_i$}: The posterior $p(\u_i|\R_{i*},\V,\lambda_u,\C_{i*})$
\begin{align*}
\propto\NM(\u_i|0,\lambda_u^{-1}\I)\prod\limits_j(\R_{ij}|\u_i^T\v_j|\C_{ij}^{-1}).
\end{align*}

\begin{figure}[!tb]
\begin{center}
\includegraphics[height=3.4cm]{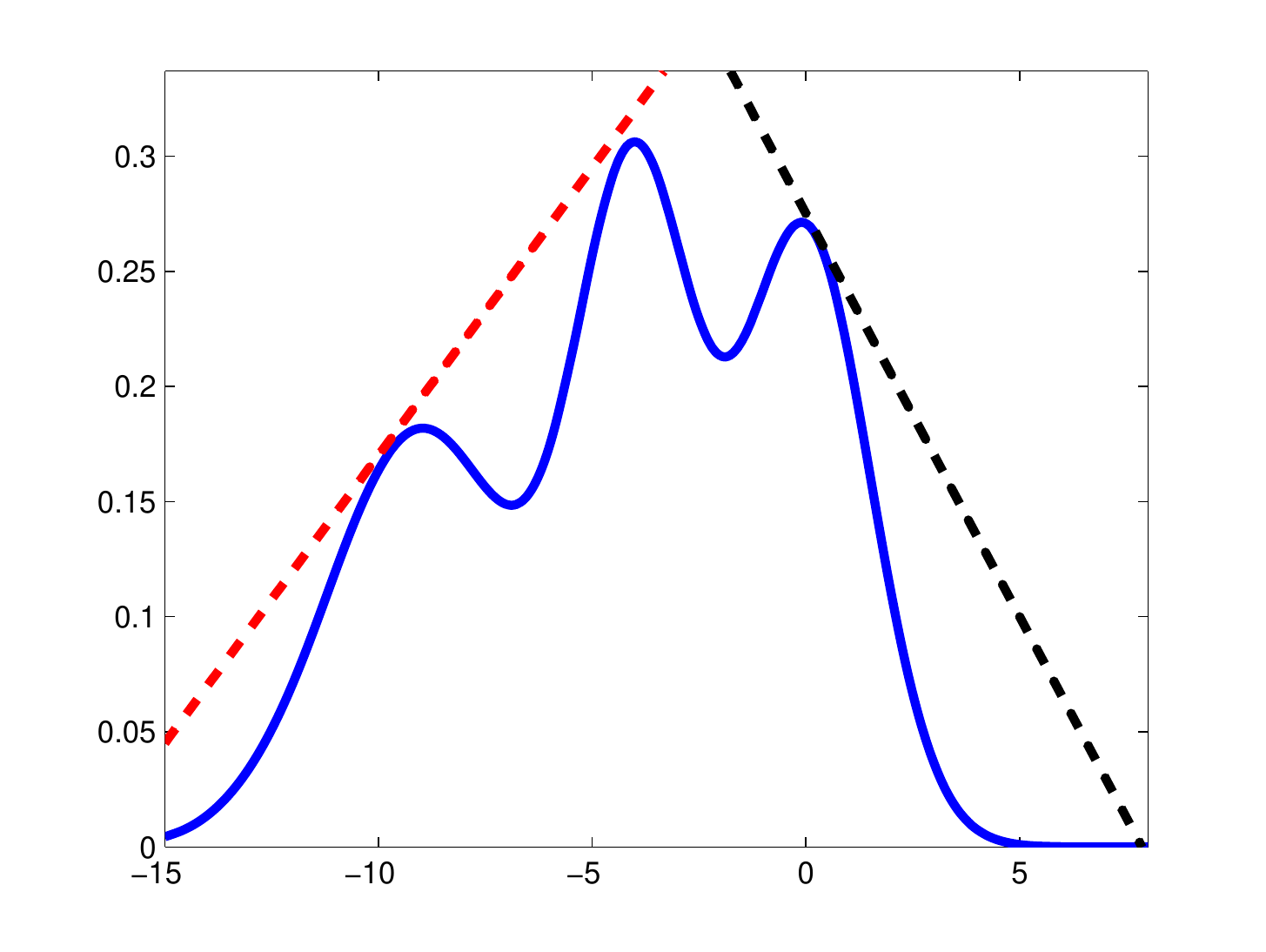}
\end{center}
\vskip -0.4in
\caption{Sampling as generalized BP.
}
\label{fig:sampling}
\vskip -0.2in
\end{figure}

Interestingly, if $\lambda_s$ goes to infinity and adaptive rejection Metropolis sampling (which involves using the gradients of the objective function to approximate the proposal distribution) is used, the sampling for $\W^+$ turns out to be a Bayesian generalized version of BP. Specifically, as \mbox{Figure \ref{fig:sampling}} shows, after getting the gradient of the loss function at one point (the red dashed line on the left), the next sample would be drawn in the region under that line, which is equivalent to a probabilistic version of BP. If a sample is above the curve of the loss function, a new tangent line (the black dashed line on the right) would be added to better approximate the distribution corresponding to the loss function. After that, samples would be drawn from the region under both lines. During the sampling, besides searching for local optima using the gradients (MAP), the algorithm also takes the variance into consideration. That is why we call it Bayesian generalized back-propagation.

\end{document}